\documentclass{article}
\usepackage[accepted]{icml2024}

\usepackage[utf8]{inputenc} %

\usepackage{graphicx}

\usepackage[T1]{fontenc}

\usepackage[hidelinks,colorlinks]{hyperref}       %
\usepackage{url}            %
\usepackage{booktabs}       %
\usepackage{amsfonts}       %
\usepackage{nicefrac}       %
\usepackage{microtype}      %
\usepackage{amssymb}
\usepackage{pifont}

\IfFileExists{headers/config/showoverfull.config}{
	\overfullrule=1cm
}{
}

\usepackage{marginnote}

\usepackage[backgroundcolor=none,linecolor=red,textsize=footnotesize]{todonotes}

\newcommand{\highlightred}[1]{\textcolor{red!70!black}{\textbf{#1}}}
\newcommand{\highlightgreen}[1]{\textcolor{green!50!black}{\textbf{#1}}}
\newcommand{\highlightyellow}[1]{\textcolor{yellow!50!black}{\textbf{#1}}}
\DeclareRobustCommand{\tool}{\textsc{ToolFuzz}\xspace}

\usepackage{etoolbox}

\newbool{includeappendix}
\setbool{includeappendix}{true} %
\IfFileExists{headers/config/noappendix.config}{
	\setbool{includeappendix}{false}
}{}

\newif\ifincludeappendixx
\ifbool{includeappendix}{
	\includeappendixxtrue
}{
	\includeappendixxfalse
}

\usepackage{xr} %
\usepackage{filecontents}

\ifbool{includeappendix}{}{
	\input{appendix-labels-loader}

	\externaldocument{appendix-labels}
}

\usepackage{acro} %

\DeclareAcronym{cli} {
    short = CLI,
    long = Command Line Interface,
}

\usepackage{listings}

\usepackage{textcomp}

\usepackage{xcolor}

\usepackage[scaled=0.8]{beramono}

\definecolor{ckeyword}{HTML}{7F0055}
\definecolor{ccomment}{HTML}{3F7F5F}
\definecolor{cstring}{HTML}{2A0099}

\lstdefinestyle{numbers}{
	numbers=left,
	framexleftmargin=20pt,
	numberstyle=\tiny,
	firstnumber=auto,
	numbersep=1em,
	xleftmargin=2em
}

\lstdefinestyle{layout}{
	frame=none,
	captionpos=b,
}

\lstdefinestyle{comment-style}{
	morecomment=[l]//,
	morecomment=[s]{/*}{*/},
	commentstyle={\color{ccomment}\itshape},
}

\lstdefinestyle{string-style}{
	morestring=[b]",%
	morestring=[b]',%
	stringstyle={\color{cstring}},
	showstringspaces=false,%
}

\lstdefinestyle{keyword-style}{
	keywordstyle={\ttfamily\bfseries},
	morekeywords={
		function,
		constructor,
		int,
		bool,
		return,
		returns,
		uint
	},
	morekeywords = [2]{},
	keywordstyle = [2]{\text},
	sensitive=true,
}

\lstdefinestyle{input-encoding}{
	inputencoding=utf8,
	extendedchars=true,
	literate=
	{ℝ}{$\reals$}1%
	{→}{$\rightarrow$}1%
	{α}{$\alpha$}1%
	{β}{$\beta$}1%
	{λ}{$\lambda$}1%
	{θ}{$\theta$}1%
	{ϕ}{$\phi$}1%
}

\lstdefinestyle{escaping}{
	moredelim={**[is][\color{blue}]{\%}{\%}},
	escapechar=|,
	mathescape=true
}

\lstdefinestyle{default-style}{
	basicstyle=\fontencoding{T1}\ttfamily\footnotesize,
	style=numbers,
	style=layout,
	style=comment-style,
	style=string-style,
	style=keyword-style,
	style=input-encoding,
	style=escaping,
	tabsize=2,
	upquote=true
}

\lstdefinelanguage{BASIC}{
	language=C++,
	style=default-style
}[keywords,comments,strings]%

\usepackage[utf8]{inputenc} %
\usepackage[T1]{fontenc}    %
\usepackage{algcompatible}
\usepackage{caption}
\usepackage{algpseudocode}
\usepackage{enumerate}
\usepackage{url}            %
\usepackage{booktabs}       %
\usepackage{amsfonts}       %
\usepackage{nicefrac}       %
\usepackage{microtype}      %
\usepackage{xcolor}         %
\usepackage[ruled,vlined]{algorithm2e}
\usepackage{tikz,booktabs,multirow,enumitem,bm}
\usepackage{colortbl}
\usepackage{tabularx}
\usepackage{subcaption}
\usepackage{wrapfig}
\usepackage{tabulary}
\usepackage{amsmath,amsthm,amsfonts}

\usepackage{amsmath,amsfonts,bm}
\usepackage{amsthm}
\usepackage{mathtools}

\def\1{\bm{1}}

\DeclareMathAlphabet{\mathsfit}{\encodingdefault}{\sfdefault}{m}{sl}
\SetMathAlphabet{\mathsfit}{bold}{\encodingdefault}{\sfdefault}{bx}{n}

\definecolor{hyperlinkblue}{HTML}{0000AA}
\hypersetup{citecolor=hyperlinkblue} %

\usepackage{tcolorbox}

\definecolor{peach}{HTML}{FFE6CC}
\definecolor{dark-peach}{HTML}{D79B00}

\definecolor{custom-gray}{HTML}{F5F5F5}
\definecolor{custom-dgray}{HTML}{666666}

\definecolor{custom-blue}{HTML}{F9FBFC}
\definecolor{custom-dblue}{HTML}{DAE8FC}

\definecolor{custom-green}{HTML}{D5E8D4}
\definecolor{custom-dgreen}{HTML}{82B366}

\definecolor{custom-red}{HTML}{F8CECC}
\definecolor{custom-dred}{HTML}{B85450}

\algnewcommand\algorithmicassert{\texttt{assert}}
\algnewcommand\Assert[1]{\State \algorithmicassert(#1)}

\usepackage[capitalize, noabbrev]{cleveref}

\crefformat{section}{\S#2#1#3}

\crefrangeformat{section}{\S#3#1#4\crefrangeconjunction\S#5#2#6}

\crefmultiformat{section}{\S#2#1#3}{\crefpairconjunction\S#2#1#3}{\crefmiddleconjunction\S#2#1#3}{\creflastconjunction\S#2#1#3}

\newcommand{\crefrangeconjunction}{--}

\crefname{listing}{Lst.}{listings}
\crefname{line}{Lin.}{Lin.}
\crefname{appendix}{App.}{App.}

\newcommand{\appref}[1]{%
	\ifbool{includeappendix}{\cref{#1}}{the appendix}%
}
\newcommand{\Appref}[1]{%
	\ifbool{includeappendix}{\cref{#1}}{The appendix}%
}

\newcommand{\OurTitle}{ToolFuzz - Automated Agent Tool Testing}

\icmltitlerunning{\OurTitle{}}

\begin{document}

\twocolumn[
\icmltitle{\OurTitle{}}

\icmlsetsymbol{equal}{*}
\begin{icmlauthorlist}
	\icmlauthor{Ivan Milev}{eth}
	\icmlauthor{Mislav Balunović}{eth}
	\icmlauthor{Maximilian Baader}{eth}
	\icmlauthor{Martin Vechev}{eth}

	imilev@ethz.ch,
	$\{$mislav.balunovic, mbaader, martin.vechev$\}$@inf.ethz.ch
	
\end{icmlauthorlist}
	
\icmlaffiliation{eth}{Department of Computer Science, ETH Zurich}
	
\icmlcorrespondingauthor{Ivan Milev}{imilev@ethz.ch}

\icmlkeywords{AI Agents, LLM Agents, Fuzzing, LLM Fuzzing, Tool Description}

\vskip 0.3in
]
\printAffiliationsAndNotice{}

\begin{abstract}
Large Language Model (LLM) Agents leverage the advanced reasoning capabilities of LLMs in real-world applications. To interface with an environment, these agents often rely on tools, such as web search or database APIs. As the agent provides the LLM with tool documentation along the user query, the completeness and correctness of this documentation is critical. However, tool documentation is often over-, under-, or ill-specified, impeding the agent's accuracy. Standard software testing approaches struggle to identify these errors as they are expressed in natural language. Thus, despite its importance, there currently exists no automated method to test the tool documentation for agents. To address this issue, we present \textbf{\tool}, the first method for automated testing of tool documentations. \tool is designed to discover two types of errors: (1) user queries leading to tool runtime errors and (2) user queries that lead to incorrect agent responses. \tool can generate a large and diverse set of natural inputs, effectively finding tool description errors at a low false positive rate. Further, we present two straightforward prompt-engineering approaches. We evaluate all three tool testing approaches on 32 common LangChain tools and 35 newly created custom tools and 2 novel benchmarks to further strengthen the assessment. We find that many publicly available tools suffer from underspecification. Specifically, we show that \tool identifies 20x more erroneous inputs compared to the prompt-engineering approaches, making it a key component for building reliable AI agents.

\end{abstract}

\section{Introduction}\label{sec:introduction}

LLM agents and compound systems~\citep{yao2023reactsynergizingreasoningacting, compound-ai-blog} aim to combine the powerful reasoning capabilities of LLMs with real-world interactions to solve complex tasks. 
An agent receives a natural language query from a user and performs a sequence of actions that interact with external \emph{tools}, such as web search~\cite{zhou2024webarenarealisticwebenvironment}, code execution~\citep{yang2024sweagentagentcomputerinterfacesenable}, or querying a database~\citep{wang2024macsqlmultiagentcollaborativeframework}, with the goal of solving the specified task. 
As such, effective interaction with these tools is critical for successful task completion.

\paragraph{Reliability Problems of Agent Tools}
To facilitate tool usage for agents, each tool is accompanied by documentation detailing its functionality and intended use, which can be provided to the agent LLM as part of the prompt. 
In practice, however, LLM agents often fail to use the available tools correctly~\citep{toolsfails2024,sweagent2024,webarenaerrors2024} because their documentation assumes human-level understanding. 
Thus, it is often underspecified (leaving crucial details implicit), overspecified (focusing narrowly on a single use case despite broader applicability), or illspecified (where the tool's functionality is not aligned with its documentation). 
Although tools play a crucial role in agent systems~\citep{yuan2024easytoolenhancingllmbasedagents,qu2024explorationmasteryenablingllms}, automated methods for detecting such errors are lacking. 
Traditional fuzzing techniques, generating seemingly meaningless random strings, fall short for agent testing, as they cannot simulate the expected natural language queries. 
For instance, when using the well-known fuzzing algorithm American Fuzzy Lop on an agent tool for several hours, it failed to find even a single example that realistically represents an LLM agent scenario.

\paragraph{Testing Agent Tools with \tool}
To address this, we introduce \tool, a novel method for automatic end-to-end tool testing.
\tool has two techniques to uncover specification errors: (1) by generating queries that lead to tool runtime errors, achieved by combining fuzzing techniques with LLM-based query generation, and (2) by generating queries that result in incorrect agent responses, using synonymous prompt generation and a series of cascading consistency and correctness checks at various stages of the agent's processing. 
Experimentally, we show that \tool successfully identifies a large number of erroneous queries, aiding in the improvement of the tool and thus its utility to the agent.

\paragraph{Main Contributions:}

\begin{itemize} 
    \item A novel, end-to-end agent-centric method for finding errors in tools, called \tool.
    \item A new benchmark suite that focuses on evaluating accurate tool utilization for file management and GitHub agents, emphasizing precise tool invocation rather than sophisticated reasoning and planning.
    \item A thorough experimental evaluation of \tool across a wide variety of agent tools and tasks.
\end{itemize}

\begin{figure*}[t]
    \centering
    \resizebox{\textwidth}{!}{\includegraphics{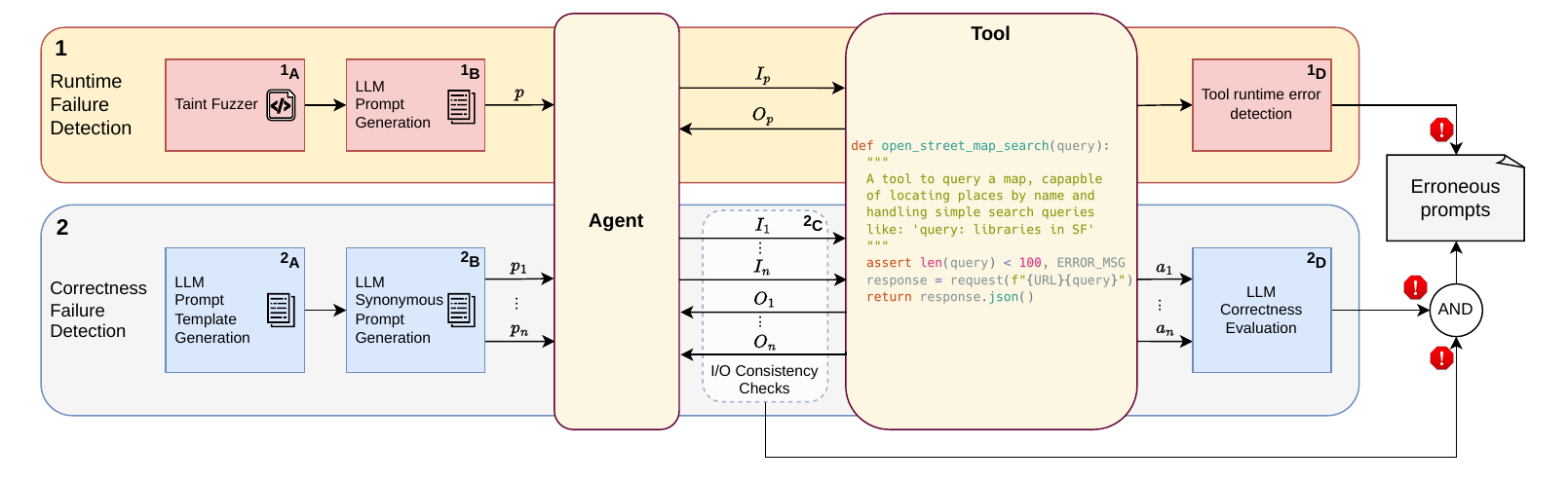}}
    \caption{
    Overview of the two error detection techniques of \tool, consisting of (1) a fuzzing based approach  and (2) an invariance based approach utilizing consistency checks. 
    Prompts are denoted by $p$ or $p_j$, tool calls by $I_p$ or $I_j$, tool responses by $O_p$ or $O_j$ and agent responses by $a$ or $a_j$. 
    }
    \label{fig:toolfuzz-overview}
\end{figure*}

\begin{figure*}[t]
    \centering
    \fontsize{7pt}{7pt}\selectfont %
    \begin{tcolorbox}[colback=peach, colframe=dark-peach, boxrule=0.3mm, width=0.9\textwidth, before skip=0mm, after skip=0.1mm]
        \begin{minipage}[t]{0.8\textwidth}
            \textbf{Tool function declaration:} def open\_street\_map\_search(query: str) $\rightarrow$ str:
        \end{minipage}%
        \begin{minipage}[t]{0.2\textwidth}
            \raggedleft\textbf{Tool} %
        \end{minipage}
        \\
        \textbf{Tool documentation:} \textit{Tool to query a map. This tool can locate places by name and simple queries such as: libraries in San Francisco. The idea of the tool is to locate places like coffee shops, offices, etc.}
    \end{tcolorbox}

    \begin{tcolorbox}[colback=custom-red, colframe=custom-dred, boxrule=0.3mm, sharp corners=south, width=0.9\textwidth, before skip=0mm, after skip=0mm]
        \begin{minipage}[t]{0.95\textwidth}
            \textbf{Fuzzing based prompt generation:}\\
            \hspace*{1em} (1) Locate places at PSC 3315, Box 8692 APO AE 21800 Unit 2602 Box 5634 DPO AE 92418 3375 Hoover Well East Andrew, ID 52728 5868 Crosby Crescent Gregoryview, AZ 14709 569 Paul Ports Apt. 406 New Saraside, TX 94798 USNS Diaz FPO AE 87563 PSC 9514, Box 8465 APO AA 91885 66299 Sullivan Village West Justinchester, WY 76520 7738 Leon Underpass Apt. 148 Clarencebury, TX 41188 00379 Stanley Roads Lake Heather, NM 52884 \\ 
            \hspace*{1em} (2) Find the nearest hospital in the Andromeda Galaxy
        \end{minipage}%
        \begin{minipage}[t]{0.05\textwidth}
            \raggedleft{$\text{\textbf{1}}_\text{\textbf{B}}$} %
        \end{minipage}
    \end{tcolorbox}
    \begin{tcolorbox}[colback=custom-blue, colframe=custom-dblue, boxrule=0.3mm, sharp corners=north, 
        top=1mm, bottom=1mm, coltitle=black, width=0.9\textwidth, 
        before skip=0mm, after skip=0.2mm]
        \begin{minipage}[t]{0.8\textwidth}
        \textbf{Synonymous prompt generation:}\\
        \hspace*{1em} (3) Find grocery store near San Jose. \\ 
        \hspace*{1em} (4) Find supermarket near SJ USA. \\
        \hspace*{1em} (5) Find food market near San Jose California. \\
        \hspace*{1em} (6) Find grocery market near SJ CA.
        \end{minipage}%
        \begin{minipage}[t]{0.2\textwidth}
            \raggedleft{$\text{\textbf{2}}_\text{\textbf{B}}$} %
        \end{minipage}
    \end{tcolorbox}

    \begin{minipage}{0.5\textwidth}
        \begin{tcolorbox}[colback=custom-red, colframe=custom-dred, boxrule=0.3mm, sharp corners=south, 
            width=\textwidth, before skip=0mm, after skip=0mm]
        \begin{minipage}[t]{0.95\textwidth}
            \textbf{Tool invocations:} \\
            \hspace*{1em}(1) open-street-map-search("query:PSC 3315, Box 8692 APO...") \\
            \hspace*{1em}(2) open-street-map-search("query:hospital in the Andromeda Galaxy")
        \end{minipage}%
        \begin{minipage}[t]{0.05\textwidth}
            \raggedleft{$\text{\textbf{1}}_\text{\textbf{invoc.}}$} %
            \end{minipage}
        \end{tcolorbox}
        \begin{tcolorbox}[colback=custom-blue, colframe=custom-dblue, boxrule=0.3mm, sharp corners=north, 
                         top=1mm, bottom=1mm, coltitle=black, width=\textwidth, 
                         before skip=0mm, after skip=0mm]
            \makebox[\textwidth][r]{$\text{\textbf{2}}_\text{\textbf{invoc.}}$} %
            \hspace*{1em}(3) open-street-map-search("query:grocery store near San Jose") \\
            \hspace*{1em}(4) open-street-map-search("query:supermarket near San Jose, USA")\\

            \hspace*{1em}(5) open-street-map-search("query:food market near San Jose California") \\
            \hspace*{1em}(6) open-street-map-search("query:grocery market near San Jose, CA")
            
        \end{tcolorbox}
    \end{minipage}%
    \hspace{0.01\textwidth} %
    \begin{minipage}{0.39\textwidth}
        \begin{tcolorbox}[colback=custom-red, colframe=custom-dred, boxrule=0.3mm, sharp corners=south, 
            width=\textwidth, before skip=0.1mm, after skip=0mm]
        \begin{minipage}[t]{0.95\textwidth}
            \textbf{Tool outputs:} \\
            \hspace*{1em}(1) \textless class \highlightred{'AssertionError'} \textgreater Query is too long... \\
            \hspace*{1em}(2) The top \highlightred{0 results} for the query are: .
        \end{minipage}%
        \begin{minipage}[t]{0.05\textwidth}
            \raggedleft{$\text{\textbf{1}}_\text{\textbf{outputs}}$} %
        \end{minipage}
        \end{tcolorbox}
        \begin{tcolorbox}[colback=custom-blue, colframe=custom-dblue, boxrule=0.3mm, sharp corners=north, top=1mm, bottom=1mm, coltitle=black, width=\textwidth,                          before skip=0mm, after skip=0mm]
            \makebox[\textwidth][r]{$\text{\textbf{2}}_\text{\textbf{outputs}}$} %
            \hspace*{1em}(3) The top \highlightred{0 results} for the query are: . \\
            \hspace*{1em}(4) The top \highlightgreen{10 results} for the query are: \highlightyellow{Safeway}...\highlightyellow{Whole Foods Market}... \\ 
            \hspace*{1em}(5) The top \highlightred{0 results} for the query are: . \\
            \hspace*{1em}(6) The top \highlightred{0 results} for the query are: . 
        \end{tcolorbox}
    \end{minipage}

    \hspace{0.02\textwidth} 
    \begin{tcolorbox}[colback=custom-green, colframe=custom-dgreen, boxrule=0.3mm, width=0.9\textwidth, before skip=-0.2mm, after skip=0mm] 
        \begin{minipage}[t]{0.8\textwidth}
            \textbf{LLM Oracle expectation:} \\
            \hspace*{1em}Grocery stores near San Jose include: \highlightyellow{Safeway}, Trader Joe's, \highlightyellow{Whole Foods Market}, and Sprouts Farmers Market.
        \end{minipage}%
        \begin{minipage}[t]{0.2\textwidth}
            \raggedleft{$\text{\textbf{2}}_\text{\textbf{D}}$} %
        \end{minipage}
    \end{tcolorbox}

    \hspace{0.02\textwidth} %
    \begin{tcolorbox}[colback=white, colframe=black, boxrule=0.3mm, width=0.9\textwidth, before skip=0.1mm, after skip=0mm]
        \begin{minipage}[t]{0.98\textwidth}
            \textbf{Agent answers:} \\
            \hspace*{1em}(1) \textless class \textbf{'AssertionError'}\textgreater Query is too long... \\
            \hspace*{1em}(2) The top \textbf{0} results for the query are: .\\
            \makebox[\textwidth]{\dotfill} \\
            \hspace*{1em}(3) It seems that I \highlightred{couldn't find any grocery stores near San Jose}. Would you like me to try a different query or look for something else? \\
            \hspace*{1em}(4) Here are some supermarkets near San Jose, USA: 1. Dai Thanh Supermarket... 2. Grocery Outlet... 5. \highlightyellow{Safeway}... 10. \highlightyellow{Whole Foods Market} \\
            \hspace*{1em}(5) It seems that \highlightred{I couldn't find any food markets near San Jose, California}. Would you like me to search for something else, such as grocery stores or farmers' markets? \\
            \hspace*{1em}(6) It seems that \highlightred{I couldn't find any grocery markets near San Jose, CA}. Would you like me to try a different search or provide information on something else?
        \end{minipage}%
        \begin{minipage}[t]{0.02\textwidth}
            \raggedleft\textbf{Agent} %
        \end{minipage}
    \end{tcolorbox}
    \vspace{1em}
    \caption{Input/Output overview for \texttt{open\_street\_map\_search} tool evaluated with \tool. Note that the numbering corresponds to the numbering of the two approaches in \cref{fig:toolfuzz-overview}. 
    }
    \label{fig:overview-example}
    \vspace{-3em}
\end{figure*}

\section{Background and Related Work}\label{sec:background}
In this section, we provide the essential background and review the most relevant prior work.

\paragraph{Language Models}
Throughout this work, we rely on generative Large Language Models (LLMs) for various tasks. 
Specifically, we use commercial instruction-tuned LLMs -- GPT-4~\citep{openai2024gpt4technicalreport}, GPT-4o, GPT-4o-mini~\citep{OpenAI}, which have shown impressive performance across various challenging benchmarks (e.g., HELM, MMLU-Pro~\citep{liang2023holisticevaluationlanguagemodels,wang2024mmluprorobustchallengingmultitask}). 
Despite this, they remain susceptible to "hallucinations" that can undermine their trustworthiness~\citep{lee2023factualityenhancedlanguagemodels, manakul2023selfcheckgptzeroresourceblackboxhallucination, mundler2024selfcontradictoryhallucinationslargelanguage}. 
To address this issue, numerous methods have emerged, employing various prompting techniques~\citep{fluri2023evaluatingsuperhumanmodelsconsistency, wang2023selfconsistencyimproveschainthought, mundler2024selfcontradictoryhallucinationslargelanguage}. 
Additionally, LLMs are used as judges~\citep{wei2023chainofthoughtpromptingelicitsreasoning, zheng2023judgingllmasajudgemtbenchchatbot} to improve accuracy. 
Furthermore, cross-checking multiple generations with assertive generations~\citep{chen2022codetcodegenerationgenerated} can help balance efficiency with more robust correctness verification.

\paragraph{LLM Agents} 
LLM agents enable LLMs to interact with external tools such as Web-APIs, databases, and code execution environments. The LLM is responsible for reasoning, planning, and tool usage to continuously enrich its context, thereby improving the quality of the final response~\citep{shinn2023reflexionlanguageagentsverbal,  wang2023voyageropenendedembodiedagent, yao2023reactsynergizingreasoningacting}. 
We define an agent system $\mathcal{A}$ as a tuple $\langle L, \mathcal{F} \rangle$, where the language model $L$ has access to a set of tools $\mathcal{F} \ni f_i$. 
As many tools are frequently wrappers around robust Web-APIs or other well-tested libraries, we focus mainly, but not only, on potential failures due to incomplete or erroneous documentation that would undermine the tools utility to the agent~\citep{qu2024explorationmasteryenablingllms, yuan2024easytoolenhancingllmbasedagents}.

\paragraph{Agent Benchmarks and Tool Usage}
Recently, there has been growing interest in benchmarking the capabilities of LLM agents on across diverse tasks such as long-horizon planning~\citep{liu2023agentbenchevaluatingllmsagents}, security analysis~\citep{debenedetti2024agentdojodynamicenvironmentevaluate, naihin2023testinglanguagemodelagents, ruan2024identifyingriskslmagents}, real-world web interactions~\citep{deng2023mind2webgeneralistagentweb, zhou2024webarenarealisticwebenvironment, yao2023webshopscalablerealworldweb} and multimodal agents~\citep{xie2024osworldbenchmarkingmultimodalagents}.
Benchmarks such as Berkeley Function Calling Leaderboard~\citep{gorillafunctioncalling}, ToolAlpaca~\citep{tang2023toolalpacageneralizedtoollearning}, ToolBench~\citep{xu2023toolmanipulationcapabilityopensource}, and MetaTool~\citep{huang2024metatoolbenchmarklargelanguage}, evaluate function-call accuracy. However, they primarily focus on the reasoning and planning aspects of the LLM rather than the correctness of the tools used and their documentation. 
While methods for enhancing tool documentation \citep{qu2024explorationmasteryenablingllms, yuan2024easytoolenhancingllmbasedagents} recognize tools and their documentation as critical components, they do not provide a methodology to identify such insufficiencies. 
Our work addresses this gap by introducing \tool, along with two custom benchmarks, a GitHub and a File Management benchmark, specifically focused on tool evaluation with an emphasis on correct tool utilization over sophisticated agent planning.

\paragraph{Fuzz Testing}
In classical software engineering, fuzzing is a widely used technique to discover errors or vulnerabilities in software by generating random or semi-random inputs~\citep{fuzzingpapers, fuzzingbook2024}. 
White-box fuzzing leverages source code insights, black-box fuzzing treats the program as a black box, and gray-box testing blends both approaches.
Recently, LLMs have been employed to enhance fuzzing in creating more semantically meaningful tests~\citep{huang2024largelanguagemodelsbased, yang2024whitefox, deng2023largelanguagemodelszeroshot}. 
However, both traditional and modern methods must be carefully managed, as a high number of false positives can undermine the credibility of the testing process and waste resources like compute and time. 
While fully LLM-driven fuzzing faces challenges in cost, scope, and reliability~\citep{jiang2024fuzzingmeetsllmschallenges}, combining fuzzing with machine learning has proven effective for testing~\citep{10.1145/3319535.3363230}. 
Consequently, we adopt a hybrid approach: instead of producing purely random strings, which would likely be out of distribution for LLMs, we combine fuzzing and LLM-based prompt generation, using partial knowledge of the tool's semantics to systematically probe agent-tool interactions.

\section{Overview}\label{sec:overview}

We now provide an overview of \tool. 
First, we explain the types of documentation errors: under-, over-, and ill-specification as well as the different types of agent failures, specifically tool runtime and correctness errors. 
Then, we describe our techniques for finding user prompts, tailored to each type of agent failure.

\paragraph{Documentation Errors}\label{sec:spec_bugs}
Several errors can arise due to faulty documentation. 
In practice, recent work has shown that agents often call tools inaccurately. 
As these calls primarily depend on the tools' documentation, it can be held accountable for a large portion of agent failures~\citep{gorillafunctioncalling, yang2024sweagentagentcomputerinterfacesenable, yao2024taubenchbenchmarktoolagentuserinteraction, ScaleAi}. 
First, documentation can be \emph{underspecified} regarding a tool's limitations, causing the LLM to use the tool in unsupported contexts, such as using \texttt{arXiv} to search for ``Italian food''. 
Additionally, underspecification can restrict tool usage, i.e. the \texttt{pubmed} tool has additional features beyond searching for paper titles which are unspecified in the documentation (see more in \Cref{sec:case_study}). 
Second, documentation can be \emph{overspecified}, artificially reducing the scope or the ways in which the tool can be used. 
Third, documentation can be \emph{illspecified}, reflecting a misalignment between a tool's functionality and its described usage. 
In practice, under- or ill-specified argument documentation can include incomplete or incorrect enumeration values, missing or outdated syntax details, or omitted relational constraints. 
These documentation errors often lead to tool misuse, highlighting the need for methods to detect these shortcomings. 

\paragraph{Agent Tool Failures}
Initial investigation reveals two distinct manifestations of these documentation errors in agent tool failures: first, runtime failures, where invalid or misformatted inputs lead to runtime tool errors, and second, correctness failures, where the tool returns incorrect results for a given user query. 
These issues often arise from a mismatch between the tool documentation, written by developers, and its interpretation by LLMs. 

\subsection{\tool}
To address the two types of agent tool failures \tool utilizes two custom techniques to detect errors: 
(1) a fuzzing inspired approach to detect runtime errors, and 
(2) an invariance-based approach to detect correctness errors. 
Both techniques of \tool are illustrated in \cref{fig:toolfuzz-overview}. 
In the following, we provide a detailed overview of these techniques using the \texttt{open-street-map-search} tool, specified in \cref{fig:overview-example}, as a running example. 
The tool processes free-form text queries to retrieve location information.

\paragraph{Runtime Failure Detection}
The runtime failure detection technique combines fuzzing with LLM-based generation to produce realistic testing prompts (see \cref{fig:toolfuzz-overview}.1). 
For fuzzing, we use our TaintFuzzer (\cref{fig:toolfuzz-overview}.$\text{1}_{\text{A}}$) to stress test the tool, collecting tool inputs that lead to runtime errors. 
In the case of the \texttt{open-street-map-search} tool, which has a length restriction on the query parameter, the fuzzer identifies a particularly long query, \textit{"PSC 3315, Box 8692 APO AE 21800 Unit 2602 Box 5634 DPO AE 92418…"}, causing a runtime error. 
This failing input is then passed to an LLM along with the tool documentation to generate a user query incorporating it. For our running example, the generated user query is \textit{``(5) Locate places at PSC…''} (\cref{fig:overview-example}.$\text{1}_{\text{B}}$).

Once the user query is generated, it is passed to the agent LLM (\cref{fig:toolfuzz-overview}.$\text{1}_{\text{B}}$), which plans a tool invocation as show in~\cref{fig:overview-example}.$\text{1}_{\text{invoc.}}$. 
The tool execution is then monitored for runtime errors. In our example, the tool raises \textit{"(5) \textless class 'Assertion Error'\textgreater Query is too long..."} (\cref{fig:overview-example}.$\text{1}_{\text{outputs}}$). 
Whenever such an error occurs, the corresponding query is collected as an erroneous prompt (\cref{fig:toolfuzz-overview}.$\text{1}_{\text{D}}$).

\paragraph{Correctness Failure Detection}
To detect correctness failures, \tool generates synonymous prompts and applies cascading checks throughout the agent process, including checks for tool argument consistency, tool output consistency, and an LLM correctness evaluation (\cref{fig:toolfuzz-overview}.2). 
First, \tool generates a template prompt, which is then populated with synonymous mask infills (\cref{fig:toolfuzz-overview}.2 A and B). 
For example, ``Find [A] near [B].'' can be instantiated with strings like ``Find grocery store near San Jose.'' and ``Find supermarket near SJ USA'' (\cref{fig:overview-example}.$\text{2}_{\text{B}}$).

Once these $n$ user queries $p_1, \dots, p_n$ are generated, they are passed to the agent LLM (\cref{fig:toolfuzz-overview}.$\text{2}_{\text{B}}$), which again plans and executes tool invocations. 
Here, \tool checks the inputs $I_1, \dots, I_n$ and outputs $O_1, \dots, O_n$ separately for consistency (\cref{fig:toolfuzz-overview}.$\text{2}_\text{C}$). 
For our running example, these consistency checks fail as not only the tool inputs $I_j$ generated by the agent LLM are not equal, but the tool outputs $O_j$ also do not match each other. 
Specifically, in \cref{fig:overview-example}.$\text{2}_{\text{invoc.}}$, we see that the input arguments do not coincide. Similarly, in \cref{fig:overview-example}.$\text{2}_{\text{outputs}}$, we see that the responses are also not equal. 

However, such checks alone may miss consistent outputs, which are obviously wrong. 
To catch these, \tool uses an LLM correctness evaluator (an LLM Oracle) to compare the agent's output against a generated expectation (\cref{fig:toolfuzz-overview}.$\text{2}_{\text{D}}$). 
In our example, the LLM Oracle expects \textit{``Grocery stores near San Jose include: Safeway...''}", but the agent instead replies \textit{``It seems that I couldn't find any grocery stores near San Jose...''} (\cref{fig:overview-example}.Agent). 
Since this response contradicts the expected answer (\cref{fig:overview-example}.$\text{2}_{\text{D}}$), \tool classifies it as incorrect.

\tool utilizes all three checks to reduce the number of false positives. The first two metrics expose description insufficiencies, while the LLM Oracle reduces the set of positives further, to a set of positives with mismatched expectations. 
The resulting set will be added to the erroneous prompts.

\section{Methodology} 
\label{sec:method}
\begin{figure*}[t]
    \centering
    \begin{lstlisting}[
        language=Python,
        linewidth=\columnwidth,
        numbers=none,
        framexleftmargin=0.1mm,
        escapechar=^
    ]
@tool('open-street-map-search')
def open_street_map_search(query: str) -> str:
    """
    A tool to query a map, capable of locating places by name and handling simple queries. 
    The queries must start with 'query: ', an example is 'query: libraries in San Francisco.' 
    The purpose of the tool is to locate places like coffee shops, offices, etc.
    """
    assert ^{\textcolor{red}{len(query)}}^ ^{\textcolor{red}{<}}^ ^{\textcolor{red}{100}}^, 'Query is too long. Query must be less than 100 characters'

    search_query = query.split('query: ')[1]

    if ^{\textcolor{red}{'supermarket'}}^ in keywords or 'coffee shop' in keywords or 'library' in keywords or 'office' in keywords:
        response = request(f"{OPEN_STREET_BUILDING_SEARCH}{query}&polygon_geojson=1&format=jsonv2")
    else:
        response = request(f"{OPEN_STREET_NATURE_SEARCH}{query}&format=jsonv2")
    response_json = response.json()
    return response_json
\end{lstlisting}
\vspace{-0.5em}
\caption{Example Implementation of the \texttt{open-street-map-search} tool.}
\label{fig:open_street_map_code}
\end{figure*}

As discussed in~\Cref{sec:spec_bugs}, we identify two main types of tool failures: \textit{runtime tool failures} and \textit{correctness failures}. We now explain the two techniques to detect these in detail. 

\subsection{Runtime tool failures}\label{sec:runtime_failure_detection}
Runtime tool failures of the agent occur when inputs break the tool under test, $f_\text{tested}$. 
To generate such inputs, we employ a two-stage process: first, a Taint Fuzzer generates inputs to trigger runtime errors for the tool in isolation (\cref{fig:toolfuzz-overview}.1). 
Then, provided that the inputs pass a sanity check against the documentation, an LLM is used to create natural user queries for evaluation on $\mathcal{A}$ (\cref{fig:toolfuzz-overview}.$\text{1}_\text{B}$). 
As the Taint Fuzzer operates independently of the tool documentation, this second step with it is crucial to minimize false positives.

\paragraph{Taint Fuzzing} 
As most tools have strong priors on the data such as syntax or semantic priors, it is essential to capture those and ensure that the fuzzer samples accordingly. 
An example of a syntax prior is given in \cref{fig:open_street_map_code}, where the input \texttt{query} is required to start with the text ``query: ''. A semantic prior could be formatted like an addresses. 
To generate inputs satisfying a specific syntax, we analyze the tool using taint object analysis and collect syntax requirements like specific JSON or CSV formats, or string splitting operations. 
This is then used by our custom generator, which generates syntax-conforming arguments based on their types and syntactic patterns. 
To generate inputs with a semantic prior, we leverage resources such as dictionaries or LMs. 
If the generated arguments lead to runtime tool errors, they are collected for future prompt generation. 
For the \texttt{open-street-map-search} tool, the fuzzer collects all generated arguments exceeding 100 characters (\cref{fig:open_street_map_code}).

\paragraph{Prompt Generation} 
While the fuzzer generates valid words, numbers, etc., these arguments may still not satisfy the requirements specified in the tool documentation. For example, \texttt{open-street-map-search} is designed and specified for finding locations, not scientific papers. 
To mitigate this issue, we conduct a sanity check against the documentation by instructing the agent's LLM to invoke the tool with the previously found arguments, given the documentation.
Refusal of the LLM indicates that the arguments do not satisfy the documented requirements. 
If the sanity check passes, a natural language prompt for $\mathcal{A}$ is generated via an LLM, based on the tool documentation and the arguments that break the tool. 
Finally, we filter for the user queries that lead to runtime tool errors when passed to the agent $\mathcal{A}$. 

\subsection{Correctness failure detection}\label{sec:correct_disc}

Unlike runtime failures, the primary challenge in detecting correctness failures is assessing the outputs of $f_\text{tested}$ in the absence of ground-truth data. 
We address this by introducing prompt sets $\mathbb{P} = \{p_1, p_2, \ldots, p_n\}$, consisting of synonymous prompts. 
The idea is that synonymous user queries will lead to synonymous agent responses given the correct tool documentation. 
Thus, a violation of this indicates faulty tool documentation. 
To check this, we employ a cascade of checks: we check whether synonymous user queries $p_1, \dots, p_n$ are (i) mapped to equivalent tool calls $I_1, \dots, I_n$, and (ii) equivalent tool outputs $O_1, \dots, O_n$ and (iii) check if the output appears plausible to an LLM. 
We find that if all these checks fail, we have sufficient evidence to treat this prompt as erroneous. 

\paragraph{Prompt Generation}
Generating the synonymous prompt set $\mathbb{P}$ involves multiple steps. 
First, an LLM ($LM_\text{prgen}$) creates a template question using the tool documentation, e.g., \textit{``Find [venue] in [city]''} (\cref{fig:open_street_map_code}). 
Utilizing the tool documentation, an LLM then generates synonymous infills for the masked words, forming $\mathbb{P}$. 
For our example, infills for \textit{``[venue]''} can include [\textit{`libraries', `public libraries', `city libraries'}], while infills for the mask \textit{``[city]''} can be [\textit{``SF'', ``San Francisco'', ``San Francisco CA''}]. 
Multiple prompts and infills are generated per tool.

\paragraph{Correctness Detection} 
Next, \tool invokes the agent $\mathcal{A}$ on $\mathbb{P}$ and collects the tool inputs $I_j$ and outputs $O_j$. 
The key insight for correctness detection is that synonymous prompts result in consistent agent responses, given the correct tool documentation. 
Thus, inconsistent tool invocations are likely to lead to inconsistent tool responses, which, in turn, will likely lead to erroneous agent responses. 

\paragraph{Input Evaluation}
Specifically, we check the \emph{input consistency} of a prompt set $\mathbb{P}$ by verifying if the values of the argument across all function inputs $I_1, \dots, I_n$ are equal to ensure identical function calls. An input consistency check failure indicates underspecification of the tool arguments.

\paragraph{Output Evaluation}
Analogously, we check the \emph{output consistency} of $\mathbb{P}$ by comparing the responses $O_1, \dots, O_n$, again via exact matching. The output consistency serves as a proxy for comparing the inherently challenging natural language agent responses. 

\paragraph{LLM Oracle}
If the agent is consistently wrong, the input output checks fail to identify the error, leading to false negatives. Similarly, for non-deterministic tools, even if all inputs are equal, the outputs might not be, resulting in false positives. 
To combat both and enhance \tool, we add a third check for plausibility. 
While each consistency check on its own is insufficient, when combined they reduce the FDR to an acceptable level:
In the absence of ground truth, we ask an LLM to answer the queries $\mathbb{P}$~\citep{mundler2024selfcontradictoryhallucinationslargelanguage}, followed by majority voting. 
The majority answer is then compared to the agent responses by an LLM Oracle, rating their similarity on a scale from 1 to 10, with 5 as the threshold~\citep{zheng2023judgingllmasajudgemtbenchchatbot}.
We note that, even though some tools require private API access, the expected answer from the majority vote is most often sufficient. 
We consider $\{p_j\}_j$ to be faulty if all checks fail simultaneously.

\section{Experimental Evaluation}\label{sec:results}

In this section, demonstrating \tool's effectiveness in error detection via evaluation on both synthetic and public tools. 
We also introduce two new agent benchmarks: one for file management tasks and another for GitHub tasks.

\subsection{Experimental setup}
We now describe the experimental setup we used, including the agent, tools, metrics, and baselines.

\paragraph{Agent}
Our method is applicable to any agentic paradigm. 
For evaluation, we focus on the widely adopted ReACT agent~\citep{yao2023reactsynergizingreasoningacting}. 
In our experiments, we employ OpenAI's GPT-4o and GPT-4o-mini models~\citep{openai2024gpt4technicalreport}, which are widely used in both research and practice. 
We conduct our experiments using LangChain~\citep{Chase_LangChain_2022} due to the large variety of community tools. 

\paragraph{Tools}\label{sec:exp:p:dataset} We curated a dataset of 32 LangChain Community tools that do not require API keys, from a total of 96 tools, to facilitate reproducibility and adoption of \tool.
Additionally, we manually developed 35 tools: 33 for runtime failure detection and 2 for correctness failure detection. More details are available in \Cref{sec:apdx:tools}.

\paragraph{Metrics}
To assess \tool's effectiveness, we manually inspect all positives (user queries generated and detected by \tool) to obtain and report the number of false positives and the False Discovery Rate (FDR). For the new File Management and GitHub agents benchmark, performance is assessed via pass rate, which is the number of successfully completed tasks over the total number of tasks.

\paragraph{Prompt Engineering Approaches}
In the absence of prior research on testing agent tools, we introduce two prompt engineering approaches based on GPT-4o~\cite{openai2024gpt4technicalreport}, each assessing tool runtime and correctness errors.
The first method called \tool \textit{gray-box testing} approaches tool testing as a gray-box scenario, where an LLM generates test prompts based solely on the tool's name and documentation. 
For runtime tool failure detection, the LLM creates prompts aimed at triggering runtime tool errors, which are then tested on the agent. For correctness failure detection, the LLM generates prompts to elicit incorrect responses from the agent, and a second LLM, the LLM judge, evaluates the agent's answers without additional context, relying only on the prompt and the agent response. 
The second method, called \tool \textit{white-box testing} mirrors the gray-box approach but additionally incorporates the tool's source code during prompt generation. 

\begin{table*}[t]
    \centering
    \caption{A comparison of runtime failure detection methods showing the number of vulnerable tools identified for specific errors. Each row represents a distinct error type, and each column corresponds to a different detection method. TF GB is \tool Gray-Box, TF WB is \tool White-Box, TF is \tool, and TF Comb. is \tool Combined.}\label{tab:failure}
    \begin{center}
        \begin{tabular}{@{} l l c c c c | l c c c c @{}}
            \toprule
            \multicolumn{2}{l}{} &
            \textbf{TF GB} &
            \textbf{TF WB} &
            \textbf{TF} &
            \textbf{TF Comb.} &
            \multicolumn{1}{l}{} & %
            \textbf{TF GB} &
            \textbf{TF WB} &
            \textbf{TF} &
            \textbf{TF Comb.}
            \\ 
            \midrule
            \multirow{6}{*}{\rotatebox[origin=c]{90}{\textbf{Synthetic tools}}} 
            & ValueError: output parsing & 1 & 1 & 0 & 1 & \multirow{6}{*}{\rotatebox[origin=c]{90}{\textbf{LangChain tools}}} 
            & 9 & 10 & 3 &\textbf{11}
            \\
            & Input grammar type error         & 4 & 2  & 2 &\textbf{6}
                                              & & 1 & 0 & 0 &1
            \\ 
            & Input grammar syntax error       & 4 & 11 & 13 &\textbf{15} 
                                            &  & 1 & 2 & 1 &2 
            \\ 
            & HTTP error                       & 2 & 1 & 5 &\textbf{5} 
                                             & & 5 & 4 & 8 &\textbf{10}
            \\ 
            & Tool specific errors             & 0 & 2 & 2 &\textbf{3} 
                                             & & 0 & 1 & 2 &\textbf{5}
            \\ 
            & Tool's output is too long        & 0 & 1 & 0 & 1
                                             & & 0 & 0 & 1 &1
            \\ 
            \bottomrule
            
        \end{tabular}
    \end{center}
\end{table*}

\paragraph{Benchmarks}
Existing benchmarks overlook the importance of tools and instead focus primarily on agent reasoning and planning. 
As \tool is the first work on agent tool testing, we introduce two new benchmarks tailored for tool evaluation. 
Specifically, these benchmarks are designed around tasks that require precise tool usage based on its documentation, rather than complex planning. 
The first benchmark, File Management Benchmark, includes 32 folder structures and 193 tasks, accompanied by scripts for setup and validation. 
The second, GitHub Benchmark, features 54 tasks within a single repository, divided into six categories, with built-in logic for task validation and environment reset. 
Both benchmarks are divided into training and test sets. 
The training set provides additional context for \tool, while the test set remains unseen by both \tool and the automatic fix pipeline.

\subsection{Evaluation on 32 LangChain Tools} \label{sec:baseline_eval}
We now present the evaluation results for \tool in detecting both runtime and correctness failures, along with its performance on the newly introduced benchmarks.

\begin{table}[t]
    \centering
    \caption{Comparison of methodologies for detecting erroneous prompts and unique runtime tool failures as well as the number of tested prompts.}\label{tab:failure_detection_all}
    \begin{center}
        \begin{tabular}{@{} l r r c @{}}
        \toprule
            \textbf{Testing} &
            \textbf{Erroneous} &
            \textbf{Tested} &
            \textbf{Unique} \\
            \textbf{method} &
            \textbf{prompts} &
            \textbf{prompts} & 
            \textbf{errors} \\ 
            \midrule %
            \tool GB & 11326 & 28824 & 38 \\
            \tool WB & 6939 & 20440 & 42 \\
            \tool              & 999 & \textbf{3297} & 41 \\
            \bottomrule %
        \end{tabular}
    \end{center}
\end{table}

\paragraph{Evaluating Runtime Failure Detection}
Runtime failure detection is performed on both our handcrafted synthetic tools dataset and the LangChain community tools~\citep{Chase_LangChain_2022}. 
Note that runtime failure detection does not suffer from false positives. 
The results are presented in \Cref{tab:failure}. 
While \tool white-box performs well on synthetic tools, as these often have explicit input validation in their source code provided to the LLM, \tool excels in identifying a more diverse set of errors, especially in tools using external APIs. 
Furthermore, \tool is significantly more effective, requiring only a fraction ($1/6$ and $1/9$) of the prompts used by the other approaches (\cref{tab:failure_detection_all}), due to \tool's efficient fuzzing process using the Faker vocabulary library~\citep{faker}. 
\tool also avoids generating degenerate prompts, offering better diversity in useful prompts (see \cref{prompt:degen} in \cref{app:baseline_prompts}).

\paragraph{Evaluating Correctness Detection}
\begin{table}[t]
    \centering
    \caption{Table comparing the number of erroneous prompts (for correctness detection) detected by different methods next to the total number of used prompts, as well as the number of true positives (TP) the number of false positives (FP) and the False Discovery  Rate (FDR).} \label{tab:correctness}
    \begin{center}
        \begin{tabular}{ @{} l c c c c c @{}}
            \toprule
            \textbf{Testing} &
            \textbf{Erroneous} &
            \textbf{Total} &
            \multirow{2}{*}{\textbf{TP}} &
            \multirow{2}{*}{\textbf{FP}} &
            \multirow{2}{*}{\textbf{FDR}}\\
            \textbf{method} &
            \textbf{prompts} &
            \textbf{prompts} & 
            & \\ 
            \midrule %
            TF GB & 124 & 7308 & 11 & 113  & 92\% \\
            TF WB & 178 & 9701 & 15 & 163 & 91\% \\
            TF    & \textbf{1156} & \textbf{2521} & 622 & 534 & \textbf{46\%} \\
            \bottomrule %
        \end{tabular}
    \end{center}
\end{table}

As discussed in \Cref{sec:correct_disc}, it is crucial to minimize false positives. The key metric is the False Discovery Rate, $\text{FDR} = \frac{\text{FP}}{\text{FP} + \text{TP}}$, where $\text{FP}$ denotes False Positives (mistakenly identified erroneous prompts), and $\text{TP}$ represents True Positives.

To distinguish false positives from true positives, we inspect them manually: 
We manually check if the agent's response matches the query. 
The results of this inspection are presented in \cref{tab:correctness}. 
It confirms that 
\tool reduces FDR by 45\% compared to our prompt-based methods, while increasing the True Positives by $20 \times$. 
\tool demonstrates superior detection capabilities due to:
\begin{enumerate}[noitemsep,topsep=0pt,partopsep=0pt,parsep=0pt]
    \item A more diverse in distribution prompt generation, compared to degenerate prompts from prompt engineering methods, especially in \tool white-box testing (\cref{prompt:degen} in \cref{app:baseline_prompts}).
    \item An assessment of input-output consistency as a strong indicator of potential correctness failures (\cref{sec:correct_disc}).
    \item The LLM Oracle's access to the \textit{expected answer} obtained via majority voting, while baseline LLMs lack context and may rely on outdated information (\cref{sec:correct_disc}).
\end{enumerate}
This enables \tool to achieve a significantly lower FDR while also being more sample efficient. 

\subsection{Evaluation on Two Custom Benchmarks}
\begin{table}
    \centering
    \caption{File Management Benchmark: Comparison of the passing test rate between different fixing methods for the \texttt{terminal} tool and the \texttt{File Management Toolkit} (FMT). Tool Description: TD, Tool source code: TSrc.}\label{tab:file_management_benchmark}
    \begin{center}
        \begin{tabular}{@{} l c c c c @{}}
        \toprule
            \multirow{2}{*}{\textbf{Tool}} & \multirow{2}{*}{\textbf{Original}} & \multicolumn{2}{c}{\textbf{GPT-4o}} & \multirow{2}{*}{\textbf{Autofix}} \\
            \cmidrule(lr){3-4}
             & & \textbf{TD} & \textbf{TD+TSrc} & \\
            \midrule %
            Terminal  & 21.53\% & 25.38\% & 19.7\% &  \textbf{30.42\%}  \\
            FMT &  22.22\% & 30.06\% & 31.9\% & \textbf{33.35\%} \\
            \bottomrule %
        \end{tabular}
    \end{center}
\end{table}

\begin{table}
    \centering
    \caption{GitHub toolkit: Comparison of the passing test rate between different fixing methods for the \texttt{GitHub} toolkit (GHT). Tool Description: TD.}\label{tab:git_toolkit}
    \begin{center}
        \begin{tabular}{@{} l c c c @{}}
        \toprule
            \textbf{Tool} & \textbf{Original} & \textbf{GPT-4o + TD} & \textbf{Autofix} \\
            \midrule %
            GHT  & 62.6\% & 64.5\% & \textbf{68.75\%}  \\
            \bottomrule %
        \end{tabular}
    \end{center}
\end{table}

We evaluated \tool's effectiveness in failure detection using ReAct agents equipped with the tool or toolkit under test on our new Github and File management benchmarks. 
For the File management benchmark, we test the \texttt{terminal} tool and the \texttt{File Management Toolkit}. 
For the github benchmark, we use the \texttt{GitHub toolkit}. 
The agent is powered by GPT-4o-mini. 

\paragraph{Automatic Documentation Fixing} \label{sec:autofix} 
Initially, we evaluate the agent's performance on the default tool documentation. We then demonstrate the utility of the erroneous prompts identified by \tool by also evaluating with LLM improved documentation.
To achieve this, the LLM receives both the erroneous examples and the original tool documentation, employing argumentative prompting~\citep{dewynter2024idlikeargumentplease}. 
We compare this informed documentation improvement with two uninformed improvements, where the LLM is provided with just the original documentation, with or without the original source code. 
Next, we discuss the agent's results on these two benchmarks and demonstrate the erroneous prompts utility. 

\paragraph{File Management Benchmark} This benchmark assesses agents equipped with either LangChain's \texttt{terminal} tool or the \texttt{File Management Toolkit} (\Cref{tab:file_management_benchmark}). 
Fixing the documentation based on our erroneous prompts improves the agent's performance in both cases by $\approx 10\%$, highlighting \tool's utility. 
The \texttt{terminal} tool showed an even greater improvement over the baseline, as its original documentation is less refined compared to the already well-crafted documentation in the toolkit. 

\paragraph{GitHub Toolkit Benchmark} 
Results for LangChain's GitHub toolkit on our custom benchmark reveal a $6\%$ improvement (\cref{tab:git_toolkit}). 
Initial runs identified critical implementation issues in the GitHub toolkit, which were resolved to allow evaluation. 
The improvements to the toolkit documentation based on \tool findings are $4\%$ higher compared to the uninformed. Similar to the File Management Toolkit, improvements are bounded due to the toolkit's already well-crafted documentation.

\subsection{Case Study} \label{sec:case_study}

We present a case study of the LangChain tool \texttt{pubmed}. Two additional case studies are available in \cref{sec:apx:case_study}. The PubMed tool has short and concise documentation, as shown in \cref{fig:case_study_tools}.A in \cref{sec:apx:case_study}. Let's now consider two of the erroneous prompts that \tool discovered:
\begin{enumerate}[noitemsep,topsep=0pt,partopsep=0pt,parsep=0pt]
    \item Can you list some papers for cancer from 2020 using RNA technology?
    \item What are the latest research findings on cancer treatment?
\end{enumerate}
Both prompts concern date handling: the first requests papers from 2020, while the second seeks the latest research. 
In the first case, \texttt{pubmed} outputs: \textit{"Published: 2024-09-15 Title: A novel small molecule ZYZ384..."}, which is clearly not from 2020. 
This error occurs because the tool invocation is: \texttt{pubmed(query="cancer RNA technology 2020")}, and this query formulation only searches in paper titles. 
The PubMed Web-API specifies a particular syntax for querying specific fields, e.g., publication date: \texttt{("2020/01/01"[Date - Publication] : "2021/01/01"[Date - Publication])}. 
However, this crucial detail is missing from the LangChain tool documentation, causing the error detected by \tool. 
We resolved this issue by updating the documentation (see \Cref{fig:case_study_tools}.E in \cref{sec:apx:case_study}). This enables the agent to now formulate an accurate query, resulting in: \texttt{pubmed(query="cancer AND RNA technology AND ("2020/01/01"[Date - Publication] : "2020/12/31"[Date - Publication])")}, which now produces correct output \texttt{``Published: 2020-11-26 Title: TUG1 long non-coding RNA...''}.

\section{Conclusion}
\label{sec:conclusion}

We introduced \tool, the first method to test tools in an agent environment. 
\tool tests systematically and automatically by building a fuzzing inspired method, integrated with LLM-based prompt generation. 
We demonstrate \tool's effectiveness in a series of experiments, uncovering more than $20 \times$ the number of erroneous prompts. 
This work opens various research directions, including the expansion of \tool to test multiple tools simultaneously to allow the discovery of cross-coordination failures or automated documentation refinement. 
\tool closes a critical gap in the testing frameworks for LLM-based agents, enabling a more robust and reliable tooling ecosystems.

\clearpage

\section*{Broader Impact}
\label{sec:broader_impact}
This paper presents work whose goal is to advance the field of Machine Learning. 
There are many potential societal consequences of our work, none which we feel must be specifically highlighted here.

\clearpage

\bibliography{main}
\bibliographystyle{unsrtnat}
\vfill
\clearpage

\message{^^JLASTREFERENCESPAGE \thepage^^J}

\ifincludeappendixx
	\newpage
	\appendix
	\onecolumn
	\section{Source code}
The source code for this work is attached as a zip file alongside it.

All the instructions on how to run the code base are written in the README.md file, located at the root directory of the project. The codebase is written in Python 3.10 using a Conda virtual environment, all dependencies are listed under the requirements.yml file -- instruction on how to setup the environment are also provided in the README.md file.

\section{Agent}\label{sec:apx:method_overview}
In this work, the main focus is on the usage of tools. This said we consider a minimal agent system design meaning just an LLM for reasoning and planning, and tools. This design choice is made to minimize system overhead. The general framework considered can be seen in \cref{fig:agent_system}.
\begin{figure}[t]
    \centering
    \includegraphics[scale=0.65]{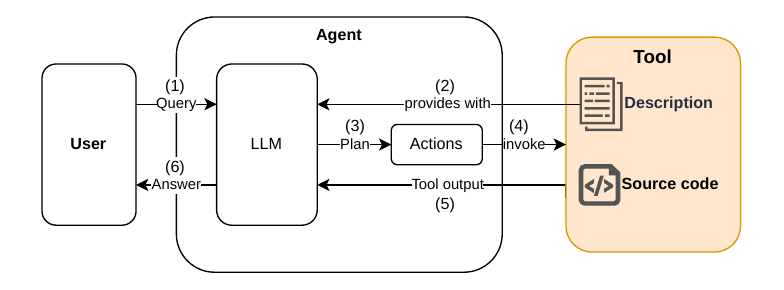}
    \caption{High-level system diagram of an LLM (AI) Agent. The flow of the diagram follows the numbering: (1) User sends a query to the Agent. Alongside with the user query the Agent is provided with the description of all tools available to it (2). With this information the LLM plans (3) actions. Some of these actions require tool invocations (4). After tool calls, an observation is made (5). Based on the observation the agent responds with an answer to the User (6).}
    \label{fig:agent_system}
\end{figure}

\section{Additional evaluation}\label{sec:apx:additional_eval}
To accompany the evaluation presented in the main article, we have conducted additional experiments and figures to further evaluate the effectiveness of \tool. 
To further evaluate the effectiveness of \tool in having low false positive rate, we have plotted the number of true and false positives per tool category in~\cref{fig:correctness_barchart}. 
The figure shows that \tool has a low false positive rate across all tool categories and the numbers shown in~\cref{tab:correctness} are not inflated by good performance on subset of tool categories.

\begin{figure}[t]
    \centering
    \includegraphics[scale=0.7]{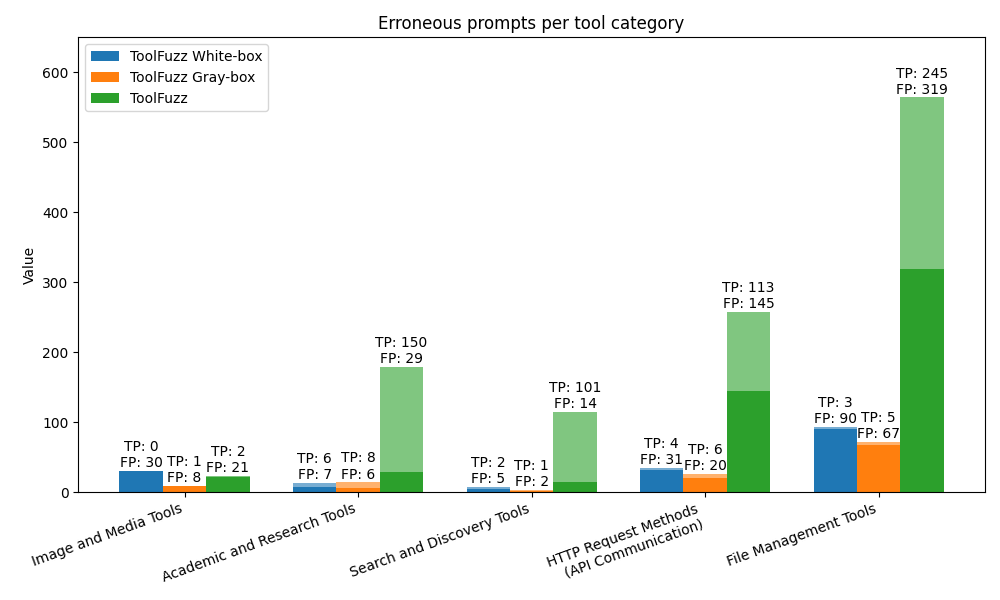}
    \caption{This bar chart illustrates the number of erroneous prompts identified by different methods across various tool categories. Each method is depicted as a stacked bar, with the upper segment showing the true positives (correctly identified erroneous prompts) and the lower segment indicating the false positives.}
    \label{fig:correctness_barchart}
\end{figure}

\subsection{Cross tool calling}
We have used prompts generated from \tool to assess whether tool documentations are too broad, leading to unintended activations across different categories. As discussed in \Cref{sec:background}, underspecification can prompt undesired tool invocations. We categorized tools by their domain and sampled prompts intended for different tool groups. Our experiment revealed that 492 prompts led to unplanned tool usages across 53 tools. Detailed results can be found in the experiment folder within the source code (./src/eval/cross\_tool\_calling). The tool categories are presented in the paragraph below.

\paragraph{Tools categorization}
As we wanted to test cross tool calling, we have grouped the tools listed in \Cref{sec:apdx:tools} into groups with respect to their domain. The groups are as follows:
\begin{enumerate}
    \item Dall-E: Dall-E-Image-Generator
    \item NASA Image and Video Library: Get Nasa Image and Video Library media metadata manifest, Get NASA Image and Video Library media metadata location, Search NASA Image and Video Library media, Get NASA Image and Video Library video captions location
    \item Knowledge Repositories: wikidata, wikipedia, stack\_exchange
    \item Academic Resources: arxiv, semanticscholar, pubmed
    \item Search engines: duckduckgosearch, duckduckgosearchresult, stack\_exchange, youtube\_search
    \item File operations: file\_search, list\_directories, terminal, python\_repl, python\_repl\_ast
    \item File deletions:  file\_delete, terminal, python\_repl, python\_repl\_ast
    \item Move files: move\_file, terminal, python\_repl, python\_repl\_ast
    \item Read files: read\_file, terminal, python\_repl, python\_repl\_ast
    \item Copy files: copy\_file, terminal, python\_repl, python\_repl\_ast
    \item HTTP Requests: request\_delete, terminal, python\_repl, python\_repl\_ast, requests\_get, requests\_patch, requests\_post, requests\_put
    \item Map search: open-street-map-search
    \item JSON Operations: json\_spec\_list\_keys, json\_spec\_get\_value
    \item Directions: open-street-map-distance
    \item GraphQL: query\_grapql
\end{enumerate}

\section{Case studies}\label{sec:apx:case_study}
\begin{figure}[t]
    \centering
    \tiny
    
    \begin{tcolorbox}[colframe=custom-dred, colback=custom-red, coltitle=black, sharp corners, width=1\textwidth]
        \makebox[\textwidth][r]{\textbf{A}}\\
        \textbf{Original tool documentation}\\
        \hspace*{1em}\textbf{Tool function declaration:} def open\_street\_map\_(query: str) $\rightarrow$ str: \\
        \hspace*{1em}\textbf{Tool documentation:} \textit{Tool to query a map. This tool can locate places by name and simple queries such as: libraries in San Francisco. The idea of the tool is to locate places like coffee shops, offices, etc.}
    \end{tcolorbox}
  
    \begin{tcolorbox}[colframe=custom-dred, colback=custom-red, coltitle=black, sharp corners, width=1\textwidth]
        \makebox[\textwidth][r]{\textbf{B}}\\
        \textbf{Original PubMed documentation}:\\
        \hspace*{1em} \textbf{Input arguments:} query - type string.\\
        \hspace*{1em} \textbf{Tool documentation:} \textit{A wrapper around PubMed. Useful for when you need to answer questions about medicine, health, and biomedical topics from biomedical literature MEDLINE, life science journals, and online books. Input should be a search query.}
    \end{tcolorbox}
    
    \begin{tcolorbox}[colframe=custom-dred, colback=custom-red, coltitle=black, sharp corners, width=1\textwidth]
        \makebox[\textwidth][r]{\textbf{C}}\\
        \textbf{Original Terminal tool documentation}:\\
        \hspace*{1em} \textbf{Input arguments:} commands - type: Union[str, List[str]], description: List of shell commands to run. Deserialized using json.loads.\\
        \hspace*{1em} \textbf{Tool documentation:} \textit{Run shell commands on this Linux machine.}
    \end{tcolorbox}
        
    \begin{tcolorbox}[colframe=custom-dgreen, colback=custom-green, coltitle=black, sharp corners, width=1\textwidth]
        \makebox[\textwidth][r]{\textbf{D}}\\
        \textbf{Fixed tool documentation}\\
        \textbf{Tool function declaration:} def open\_street\_map\_(query: str) $\rightarrow$ str: \\
        \textbf{Tool documentation:} \textit{Tool which can find a route between two locations and give back the distance in km of that route. The route is on roads that can be driven with a car. The tool provides route distance in km for a car trip between the two locations. The two locations can be cities or concrete places i.e. office buildings, shops, parks, and so on. \textbf{Tool arguments with cities must always include the full names and countries i.e. NYC $\rightarrow$ New York City, USA or Paris $\rightarrow$ Paris, France.}}
    \end{tcolorbox}
        
    \begin{tcolorbox}[colframe=custom-dgreen, colback=custom-green, coltitle=black, sharp corners, width=1\textwidth]
        \makebox[\textwidth][r]{\textbf{E}}\\
        \textbf{Fixed PubMed documentation}:\\
        \hspace*{1em} \textbf{Input arguments:} query - type string.\\
        \hspace*{1em} \textbf{Tool documentation:} \textit{A wrapper around PubMed. Useful for when you need to answer questions about medicine, health, and biomedical topics from biomedical literature, MEDLINE, life science journals, and online books. Input should be a search query. \textbf{The query has special syntax for different fields in the paper:\\
    The list of available search fields is: All Fields, Author, Date - Create, Date - Publication, EC/RN Number, Editor, Title, Title/Abstract, Transliterated Title, VolumeFor some fields i.e. the date fields ranges are available. Here is an example of a date query and its usage.\\
    List papers that are from 2010: ("2010/01/01"[Date - Entry] : "2011/01/01"[Date - Entry])\\
    Most of the other fields are used as follows:\\
    List papers with first author John Doe: John Doe[Author].\\
    In order to combine a filter with multiple fields we use AND. Here is an example:\\
    What are the papers with Last Author Ivan from 2020 until now:(Ivan[Author - Last]) AND (("2020/02"[Date - Create] : "3000"[Date - Create]))}}

    \end{tcolorbox}
        
    \begin{tcolorbox}[colframe=custom-dgreen, colback=custom-green, coltitle=black, sharp corners, width=1\textwidth]
        \makebox[\textwidth][r]{\textbf{F}}\\
        \textbf{Fixed Terminal tool documentation}:\\
        \hspace*{1em} \textbf{Input arguments:} commands - type: Union[str, List[str]], description: List of shell commands to run. Deserialized using json.loads.\\
        \hspace*{1em} \textbf{Tool documentation:} \textit{Execute shell commands on this Linux machine. \textbf{Use clear, concise commands with relative paths. Enclose paths with spaces in quotes. Validate command syntax and file paths. The tool can be invoked multiple times to ensure correct file system state.}}
    \end{tcolorbox}
    
    \caption{ a case study comparing the original and improved descriptions of three tools: \texttt{open\_street\_map}, \texttt{pubmed}, and \texttt{terminal}. The first set of three red boxes represents the original descriptions of these tools, which are prone to failures, while the second set of three green boxes shows the same tools with improved descriptions designed to reduce errors.}
    \label{fig:case_study_tools}
\end{figure}

In this section, we present two additional case studies in addition to the one presented in the main article (\cref{sec:case_study}). The tools we are looking at are \texttt{open-street-map-directions}, tool for navigating on open street map which is developed by the authors and a LangChain tool \texttt{terminal} which is designed to execute shell commands on the host system.

\paragraph{OpenStreetMap directions}
Now let's look at an example of the open street map. Here are the prompts in the prompt set $\mathbb{P}$:\\
\textcolor{blue!70}{1. What is the car route distance from Paris to Lyon?}\\
\textcolor{blue!70}{2. What is the car route distance from Paris, France to Lyon, France?}\\
\textcolor{orange!70}{3. What is the car route distance from City of Paris to City of Lyon?}\\
\textcolor{blue!70}{4. What is the car route distance from Paris FR to Lyon FR?}\\
\textcolor{blue!70}{5. What is the car route distance from Paris France to Lyon France?}

For this prompt set the generated expected answer is: \textit{"The road distance between Paris and Lyon is approximately 465 kilometers (289 miles)."}.

After invoking the agent with those prompts, there are 2 output buckets. The first bucket contains: \textcolor{blue!70}{1, 2, 4 and 5} with tool output -- \textcolor{blue!70}{\textit{"The distance between City of Paris to Lyon is 939224.1 meters"}}, while the second is just one prompt \textcolor{orange!70}{3} with output: \textcolor{orange!70}{\textit{"The distance between Paris to Lyon is 465460.3 meters"}}. 

The input argument buckets are three, grouped as follows:
\begin{itemize}
    \item bucket 1: \{1\} with parameters:\\ \texttt{\{from\_location\_query: "Paris", to\_location\_query: "Lyon"\}}
    \item bucket 2: \{2, 4, 5\} with parameters:\\ \texttt{\{from\_location\_query: "Paris, France", to\_location\_query: "Lyon, France"\}} 
    \item bucket 3: \{3\} with input parameters:\\ \texttt{\{from\_location\_query: "City of Paris", to\_location\_query: "City of Lyon"\}}
\end{itemize}

From the bucketing, we can make the observation that the inputs with full country name work well, and using just a city name works half the time (input bucket 1 and bucket 3) so one way to fix this is to add the requirement of country name in the argument to the tool's description. The fixed description can be found in \cref{fig:case_study_tools}.D. With the fixed description, the example prompt set passes.

\paragraph{Terminal} The terminal tool has a very short and generic description as shown in~\cref{fig:case_study_tools}.C. This description, however, does not explain at all how to properly use the tool i.e. in what way the command parameters have to be presented, which commands can be used etc.

Here is an erroneous set of prompts found by \tool:
\begin{enumerate}
    \item How big is the assignment\_1\_draft.docx file in the Contract Law assignments folder?
    \item Find the size of the draft of assignment 1 file in the \\ /[FULL\_PATH]/Law\_Student\_Files/Academics/Courses/Semester\_1/... directory.
    \item What is the file size of the first draft of assignment 1 in the Contract Law assignments folder?
    \item Find the size of the assignment one draft file in the \\ /[FULL\_PATH]/Law\_Student\_Files/Academics/Courses/Semester\_1/... directory.
\end{enumerate}

The FULL\_PATH is in fact the full path; however, we have masked it out as it is part of the machine's file structure.

The prompt set has the goal of finding the size of the first Assignment in the Contract law assignments folder. It is important to note that the agent is unaware of the directory in which it is running.

In this case, both the number of input and output buckets is four -- each prompt resulted in a unique input parameters and tool output:
\begin{enumerate}
    \item tool input: \texttt{\{"commands":"du -h \\ ~/Contract Law/assignments/assignment\_1\_draft.docx"\}} \\ tool output: \texttt{cannot access \\ '/Contract Law/assignments/assignment\_1\_draft.docx': No such file or directory}
    \item tool input: \\ \texttt{\{"commands":"du -sh /[FULL\_PATH]/Assignments/assignment\_1\_draft.*"\}} \\ tool output: \texttt{0 /[FULL\_PATH]/assignment\_1\_draft.docx}
    \item tool input: \texttt{\{"commands":"ls -lh \textasciitilde/Contract Law/assignments/ | grep 'assignment 1 draft'"\}} \\ tool output: "\texttt{ls: cannot access '\textasciitilde/Contract Law/': No such file or directory}"
    \item tool input: \texttt{\{"commands":"ls [FULL\_PATH]/Law\_Student\_Files/Templates"\}} \\ tool output: ""
\end{enumerate}

As can be seen, each prompt results in unique input parameters and tool output. We can see that twice in prompts 1 and 3 the tool tries to access invalid paths. For prompt 4 the path is correct; however, the task at hand is not just to list a directory. The general error here seems to be related to the provided paths -- the description does not give any inside information to the agent on where the Terminal tool is or how to navigate. With this information found by \tool and using the previously described autofix method (\cref{sec:autofix}) a new description is generated (\cref{fig:case_study_tools}). With this new description, the prompt set is now valid by just using ls multiple times to properly navigate the file system.

\section{\tool Prompts} \label{app:tool_prompts}
In this subsection of the Appendix all prompts used for \tool are listed. For all the following prompts, string templates are used -- \texttt{\{variable\}} will be inserted with the corresponding variable when the prompt is constructed.

\subsection{\tool Runtime tool failure detection}
\begin{figure}[t]
    \centering
    \tiny
    \begin{minipage}{\textwidth} %
        \fbox{ 
        \parbox{\textwidth}{
            Please invoke the \{tool.name\} exactly like: \{tool.name\}(\{bad\_arg\})
        }}
    \end{minipage}
    \caption{Sanity check prompt for forcing the agent to invoke the tool under test with the erroneous arguments.}
    \label{prompt:sanity_check}
\end{figure}

\begin{figure}[t]
    \centering
    \tiny
    \begin{minipage}{\textwidth} %
        \fbox{ 
            \parbox{\textwidth}{
Imagine you are a user who is using an AI agent. You have the following agent with its tool at your disposal:\\\{tool\_prompt\}\\
\{format\_instructions\}\\
Come up with prompts which will invoke the tool with one of these predefined arguments: \{bad\_args\}. Make use of the given arguments!
            }
        }
    \end{minipage}
    \caption{Prompt for generating realistic user prompt which will invoke the tool under test with erronoeous arguments.}
    \label{prompt:args_to_prompt}
\end{figure}

For the runtime error detection as explained in \cref{sec:method} the heavy lifting is done by the Fuzzer so an LLM is leveraged only for sanity check with the prompt \cref{prompt:sanity_check} and for converting the erroneous argument into a user query which is done with \cref{prompt:args_to_prompt}.

\subsection{\tool correctness detection}
\paragraph{Prompt set generation}
\begin{figure}[t]
    \centering
    \tiny
    \begin{minipage}{\textwidth} %
        \fbox{ 
        \parbox{\textwidth}{
        Can you generate template questions according to a tool description. Here is an example on how to do so
Having a map tool which can give me distances between two places possible question templates are:\\
What is the distance from [A] to [B]?\\
How much time would it take to go from [A] to [B]?\\
If I start from [A] and go to [B] with [C] km/h average speed how much time would it take me?\\

For tools that are knowledge based i.e. news, fact stores etc. You can generate questions similar to:\\
What do you know about [A]?\\
What is/are [A] for [B]?\\
Is it true that [A] is [B]?\\
Is [A] related to [B]? \\
In what year did [A] happen? \\
When was [A] born? \\
Find [A] in [B]?

[A] my work to [B].

[A] from [B] to [C].

Also some more specific questions like: \\
Is it true that Mr [A] was related to Mrs. [B]? \\
I am at [A], how much time it will take me to go from the closest [B] to the [C] airport? \\
Find an article/paper/document written by [A] on topic [B]? \\

Now having a tool with description: '\{tool\_prompt\}'\\
And the tool context: \{tool\_context\} \\

Please generate templates, try to generate more specific templates.
\{format\_instructions\}
        }}
    \end{minipage}
    \caption{Prompt for generating template questions/prompts for the agent, given the tool under test description and additional context if needed.}
    \label{prompt:question_template}
\end{figure}

\begin{figure}[t]
    \centering
    \tiny
    \begin{minipage}{\textwidth} %
        \fbox{ 
        \parbox{\textwidth}{
Imagine you are a user who is using an AI agent and just ask it in normal speech. 

Please generate appropriate template input values for the given template:

'\{template\_prompt\}'

They should be related to the context of tool:

\{tool\_prompt\}

And the tool context: \{tool\_context\}

These inputs should be synonyms or different way of expressing the same thing.

Here is an example:

Template: 'What are some [A] in [B]?'

Infills for A: ['Coffee Shop', 'Cafeteria', 'Coffeehouse', 'Café']

Infills for B: ['Zurich', 'ZH', 'Zurich Switzerland', 'ZH CH', 'ZH Switzerland', 'Zurich CH']

Template: 'Who is [A]?'

Infills for A: ['Albert Einstein', 'A. Einstein', 'Alb. Einstein', 'Einstein']

Template: 'When did [A] happen?'

Infills for A: ['World War I' , 'World War One', 'WW 1', 'First World War']

Template: 'What are the latest news in [A]?

Infills for A: ['cinema', 'hollywood', 'kino', 'movies', 'show business']

Template: 'What are the [A] in [B]?'

Infills for A: ['latest news', 'current events', 'breaking news', 'daily news', 'daily events']

Infills for B: ['politics', 'government', 'public affairs']

Template: 'I am at [A], how much time it will take me to go from the closest [B] to the [C]?'

Infills for A: ['Zurich HB', 'Zurich main train station', 'Zurich main station']

Infills for B: ['Mc Donalds', 'fast food restaurant McDonalds', 'McD burgers']

Infills for C: ['ETH HG Bibliothek', 'ETH main building library', 'ETH main library']

Template: 'Can you find [A] in [B]?'

Infills for A: ['family picture', 'png with the family', 'family photo', 'family portrait']

Infills for B: ['the home directory', 'my workspace', 'main directory']

Template: '[A] [B] to [C]'

Infills for A: ['Submit', 'Send', 'Upload', 'Commit']

Infills for B: ['main.py', 'the main python file', 'src/main', 'the main source file']

Infills for C: ['the server', 'the cloud', 'the repository', 'the remote branch']

Template: '[A] my work to [B]'

Infills for A: ['Move', 'Transfer', 'Cut']

Infills for B: ['archive folder', 'the archive']

Please DO NOT use any of the already generated examples: \{used\_args\}.

\{format\_instructions\}.
        
        }}
    \end{minipage}
    \caption{Prompt for generating infills for the masks in the prompt templates.}
    \label{prompt:infills}
\end{figure}

\begin{figure}[t]
    \centering
    \tiny
    \begin{minipage}{\textwidth} %
        \fbox{ 
        \parbox{\textwidth}{
Given the following tool description: '\{tool\_prompt\}' and the following tool prompts that are synonymous: '\{prompts\}'
Please make such that the prompts are like a person would write it and not a machine, so nothing too concrete but also not too vague.

\{format\_instructions\}
        }}
    \end{minipage}
    \caption{Prompt for making more human like prompts. In some cases prompts are too specific i.e. full paths or full identificators which users might shorten or write more intuitively.}
    \label{prompt:humanize}
\end{figure}

The generation of prompt sets as described in \Cref{sec:correct_disc} involves multiple LLM generations. Firstly, the prompt template is generated given the tool description and in some cases additional context, the full prompt is given in \cref{prompt:question_template}

The next step is to infill the generated template questions/prompts with synonymous phrases. For this, the following prompt is used to generate the phrases which are later inserted in the string templates \cref{prompt:infills}.

As in some cases when giving context to the prompt generation, some facts are taken directly from the context which is not very human, so we have also introduced a humanizing prompt \cref{prompt:humanize}.

\subsection{LLM Oracle}
\begin{figure}[t]
    \centering
    \tiny
    \begin{minipage}{\textwidth} %
        \fbox{ 
        \parbox{\textwidth}{
You are emulating the following tool: \{tool\_prompt\}. Given the tool return value for the following questions:
\{questions\}
\\
Example:

Tool description: Tool which can find a route between two locations and give back the distance in km of that route. The route is on rodes that can be driven with car. The tool provides route distance in km for car trip between the two locations.

The two locations can be cities or concrete places i.e. office buildings, shops, parks and so on.
\\

Questions:

What is the distance between Sofia and Zurich?

What is the distance between SF and ZH?

What is the distance between Sofia BG and Zurich CH?

What is the distance between Sofia Bulgaria and Zurich Switzerland?
\\

Answers:

The road distance between Sofia, Bulgaria, and Zurich, Switzerland is approximately 1,450 kilometers (900 miles).

If "SF" refers to San Francisco (SF), California, USA and "ZH" refers to Zurich (ZH), Switzerland, the distance is vast. Since it's impossible to drive directly due to the Atlantic Ocean, the driving distance would be irrelevant. However, hypothetically driving the distance across land would exceed 9,500 kilometers (5,900 miles).

"BG" stands for Bulgaria and "CH" stands for Switzerland. The road distance between Sofia, Bulgaria, and Zurich, Switzerland is approximately 1,450 kilometers (900 miles)

The road distance between Sofia, Bulgaria, and Zurich, Switzerland is about 1,450 kilometers (900 miles).

\{format\_instructions\}.
        }}
    \end{minipage}
    \caption{Prompt for generating answers for given question. Used for expectation generation for the LLM Oracle.}
    \label{prompt:llm_answer}
\end{figure}

\begin{figure}[t]
    \centering
    \tiny
    \begin{minipage}{\textwidth} %
        \fbox{ 
        \parbox{\textwidth}{
You have the following sentences:
\{sentences\}

Make once sentence which is factually correct to you and most of the sentences, not required to represent all of the answer but MUST be approximately correct.
\\

Examples:
\\

Sentences:

The road distance between Sofia, Bulgaria, and Zurich, Switzerland is approximately 1,450 kilometers (900 miles).

If "SF" refers to San Francisco (SF), California, USA and "ZH" refers to Zurich (ZH), Switzerland, the distance is vast. Since it's impossible to drive directly due to the Atlantic Ocean, the driving distance would be irrelevant. However, hypothetically driving the distance across land would exceed 9,500 kilometers (5,900 miles).

"BG" stands for Bulgaria and "CH" stands for Switzerland. The road distance between Sofia, Bulgaria, and Zurich, Switzerland is approximately 1,450 kilometers (900 miles)

The road distance between Sofia, Bulgaria, and Zurich, Switzerland is about 1,450 kilometers (900 miles).

Answer: The expected distance is around 1500 kilometers.
\{format\_instructions\}.
}}
    \end{minipage}
    \caption{Prompt for generating the LLM Expectation for particular tool set $\mathbb{P}$}
    \label{prompt:llm_expectation}
\end{figure}

\begin{figure}[t]
    \centering
    \tiny
    \begin{minipage}{\textwidth} %
        \fbox{ 
        \parbox{\textwidth}{
You are assessing an agent RESPONSE. You should give a score between 1 and 10. 10 means that the answer satisfies both criteria: first is the tool output and second is LLM expectation. The agent response should be within 10\% margin of error or false otherwise. 0 is for when both criteria are not satisfied.\\
Alongside the evaluation of true/false, you should provide reasoning for it.\\
\\
Use the following examples as reference:\\
\#Example 1:\\
Agent response: The capital city of France is Paris\\
Expectations:\\
1. No results found for the query\\
2. Paris\\
correctness\_degree: 5\\
Reason: The response doesn't comply with the first expectation: 'No results found for the query'\\
\\
\#Example 2:\\
Agent response: As of now, the current survey results for the 2024 US presidential election show Kamala Harris leading Donald Trump. Here are the key figures:\\
\\
- **Kamala Harris**: 46\%\\
- **Donald Trump**: 48\%\\
- **Robert F. Kennedy Jr.**: 5.5\%\\
\\
Harris's popularity has seen a significant increase, with half of voters viewing her favorably, which is an improvement from 43\% the previous week.\\
Expectations:\\
1. Kamala Harris, Donald Trump.\\
2. The current survey results for the 2024 US presidential election show candidate A leading with X\%, followed by candidate B with Y\%, and other candidates with Z\% and W\%.\\
correctness\_degree: 9\\
Reason: The response mentions both Kamala Harris and Donald Trump, thus complying with the first expectation, and also presents percentages, thus complying with the second.\\
\\
\#Example 3:\\
Agent response: If you're looking for a solid, mid-range option, expect to spend around \$400 to \$500, there is one cheaper option around 350 USD\\
Expectation:\\
1. \$347.99\\
2. A TV sells in the range of 400-800 USD\\
correctness\_degree: 8\\
Reason: The response is close enough to the expected values and is based on the expected tool output.\\
\\
\#Example 4:\\
Agent response: The distance between LA and Sofia is 1,047 miles\\
Expectation:\\
1. Distance from Sofia to LA is 10,633 km\\
2. The distance between Sofia and LA is 10,000 KM\\
correctness\_degree: 2\\
Reason: The response doesn't comply with both expectations - the error is too big, but is still mentioning the correct names.\\
\\
Now please evaluate the following agent response and give the reason for your evaluation:\\
'\{agent\_output\}'\\
Expectations:\\
1. \{tool\_output\}\\
2. \{expected\}\\
\{format\_instructions\}\\
        }}
    \end{minipage}
    \caption{Prompt used by the LLM Oracle to evaluate the correctness of an agent answer given the tool and expectation as well as the output of the tool under test}
    \label{prompt:llm_eval}
\end{figure}

The LLM Oracle has to evaluate the Agent answer according to an expectation. Firstly, the expectation is generated.

The LLM expectation generation is again a multistage process. Firstly, an LLM is prompted to answer to each of the generated prompts from a set of prompts $p \in \mathbb{P}$ -- \cref{prompt:llm_answer}. After all the answers are generated, we use them to prompt again an LLM to come up with the LLM expectation -- \cref{prompt:llm_expectation}.

Lastly, the LLM oracle has to evaluate the Agent output, for this prompt we are using both reasoning and scoring between 1 and 10 as previous experiments showed that making the LLM to just evaluate with yes/no gave more false positives -- \cref{prompt:llm_eval}.

\subsection{Baseline prompts} \label{app:baseline_prompts}

\begin{figure}[t]
    \centering
    \tiny
    \begin{minipage}{\textwidth} %
        \fbox{ 
        \parbox{\textwidth}{
        You are an AI assistant tester. The idea is to come up with prompts which will make the following tool crash.\\
        Example:\\
        Tool information: This tool can divide two numbers.\\
        Crash prompts: [\\
        "Can you please divide 10 by 0",\\
        "If I have no friends left how can I divide my apples evenly between them?"\\
        ]\\
        \\
        Tool information: \{tool\_info\}\\
        Now generate the crash prompts, you MUST give the full string prompt and no further logic has to be run on these prompts. I.e., NO "a" + "b" or other similar logic.\\
        \{format\_instructions\}\\
        }}
    \end{minipage}
    \caption{Prompt for generating prompts with the intention of triggering runtime tool failures.}
    \label{prompt:baseline_crash}
\end{figure}

\begin{figure}[t]
    \centering
    \tiny
    \begin{minipage}{\textwidth} %
        \fbox{ 
        \parbox{\textwidth}{
        You are an AI assistant tester. The idea is to come up with prompts which will make the following tool produce an incorrect answer.\\
        Tool information: \{tool\_info\}\\
        \\
        Please generate prompts which will make the following tool produce an incorrect answer.\\
        \{format\_instructions\}\\
        }}
    \end{minipage}
    \caption{Prompt for generating prompts for correctness evaluation.}
    \label{prompt:baseline_correctness}
\end{figure}

\begin{figure}[t]
    \centering
    \tiny
    \begin{minipage}{\textwidth} %
        \fbox{ 
        \parbox{\textwidth}{
        You are an AI assistant judge. Your job is to evaluate and reason on the correctness of the given answer knowing the question is:\\
        \{question\}\\
        The provided answer is: \{answer\}\\
        \\
        Is the answer correct? Please provide reasoning.\\
        \{format\_instructions\}\\
        }}
    \end{minipage}
    \caption{The prompt used by the LLM Judge for baseline correctness evaluation}
    \label{prompt:baseline_eval}
\end{figure}

\begin{figure}[t]
    \centering
    \tiny
    \begin{minipage}{\textwidth} %
        \fbox{%
            \parbox{\textwidth}{%
                \textbf{Degenerate prompt generations}\\
                \hspace*{1em}\texttt{'/home/user/../../../../../../../../../../../../../../../../../../../../../../../../../../../../../../../../../../../../../../../..'}\\
                \hspace*{1em}\texttt{Search for pattern: '*' dir\_path: '/var/lib/dpkg/lock-frontend-frontend-frontend-frontend-frontend-frontend-frontend-...'}\\
                \textbf{Out of distribution examples:}\\
                \hspace*{1em}Search for the latest recipes in Italian cuisine on arxiv.\\
                \hspace*{1em}Search for the most popular movies of all time on arxiv.\\
                \hspace*{1em}Search for the Garden of Eden.\\ 
                \hspace*{1em}Find the nearest time machine rental.\\ 
                \hspace*{1em}Find the nearest Hogwarts School of Witchcraft and Wizardry.\\
                \hspace*{1em}Find the nearest black hole.\\
                \hspace*{1em}Find the population of New York.\\
                \hspace*{1em}Search for the nearest alien spaceship.
            }%
        }%
    \end{minipage}
    \caption{Out of distribution and degenerate prompt generation by the baselines}
    \label{prompt:degen}
\end{figure}

For the baselines from \Cref{sec:baseline_eval}. We use the following prompt to generate test prompts - \cref{prompt:baseline_crash} and \cref{prompt:baseline_correctness}. Here, the main difference is that in the white box scenario the \texttt{tool\_info} variable will contain both the tool description and the tool source code while in the gray box the tool source code is not included.

In addition to these prompts, the baseline also has their own LLM judge using the following prompt for evaluation -- \cref{prompt:baseline_eval}.

As mentioned in \Cref{sec:results} the baselines can fall into degenerate string generation - examples of such cases can be seen in \cref{prompt:degen}.

\section{Benchmarks}
\subsection{File Management toolkit benchmark}\label{app:benchmarks:file_management}

\begin{figure}[t]
    \centering
    \includegraphics[scale=0.7]{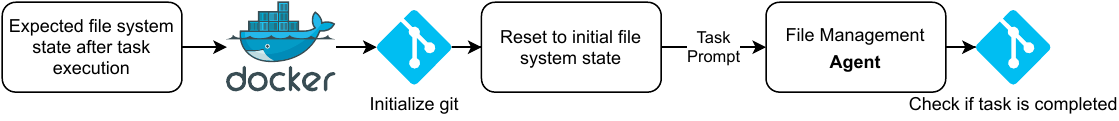}
    \caption{File management benchmark setup}
    \label{fig:file_management_diagram}
\end{figure}

\begin{figure}[t]
    \centering
    \tiny
    \begin{minipage}{\textwidth} %
        \fbox{ 
            \parbox{\textwidth}{
You are an AI Agent Tool developer can you make the description of a tool more precise and clear. The problem is that the LLM sometimes doesn’t translate the queries correctly.\\
This comes from the fact that the tool description doesn’t explain what exactly is allowed and how to use the tool correctly.\\
The current description of the tool is: \{tool\_description\}\\

Here are some failing examples of the tool in action:\\
\{bad\_examples\}\\

Given the bad examples please identify the main issues on those examples and what is the cause of that issue.\\
How can these issues be avoided by validation i.e. with this tool or external resources or it’s just a user mistake? List the main issues and how to avoid them.\\

Now that the issues and how to avoid them are clear. Please create tool description that addresses these issues. The description is a manual on how to use the tool correctly and what is allowed and what is not.\\
It should explain how to avoid the issues that were found in the examples. Add that the tool can be invoked multiple times for better validation of the file system state.\\

Also give few examples if you think they are applicable.\\

Please provide description which reflects these issues. The new description shouldn’t be longer than 100 words.\\

\{format\_instructions\}
            }}
    \end{minipage}
    \caption{Prompt for automatic tool description fixing, based on the tool's description and as well as a set of prompts resulting into tool failures (bad\_examples) found by \tool.}
    \label{prompt:autofix_us}
\end{figure}

\begin{figure}[t]
    \centering
    \tiny
    \begin{minipage}{\textwidth} %
        \fbox{ 
            \parbox{\textwidth}{
You are an AI Agent Tool developer can you make the description of a tool more precise and clear. The problem is that the LLM sometimes doesn\'t translate the queries correctly.\\
This comes from the fact that the tool description doesn\'t explain what exactly is allowed and how to use the tool correctly.\\
The current description of the tool is: \{tool\_description\}\\

The description has to be a manual on how to use the tool correctly and what is allowed and what is not.\\
Please provide just the new description of the tool.\\
\{format\_instructions\}
            }
        }
    \end{minipage}
    \caption{Prompt for automatic tool description fixing, based only on the tool description.}
    \label{prompt:autofix_desc}
\end{figure}

\begin{figure}[t]
    \centering
    \tiny
    \begin{minipage}{\textwidth} %
        \fbox{ 
            \parbox{\textwidth}{
You are an AI Agent Tool developer can you make the description of a tool more precise and clear. The problem is that the LLM sometimes doesn\'t translate the queries correctly.\\
This comes from the fact that the tool description doesn\'t explain what exactly is allowed and how to use the tool correctly.\\
The current description of the tool is: \{tool\_description\}\\

Here are some failing examples of the tool in action:\\
\{bad\_examples\}\\

Given the bad examples please identify the main issues on those examples and what is the cause of that issue.\\
How can these issues be avoided by validation i.e. with this tool or external resources or it\'s just a user mistake? List the main issues and how to avoid them.\\

Now that the issues and how to avoid them are clear. Please create tool description that addresses these issues. The description is a manual on how to use the tool correctly and what is allowed and what is not.\\
It should explain how to avoid the issues that were found in the examples. Add that the tool can be invoked multiple times for better validation of the file system state.\\

Also give few examples if you think they are applicable.\\

Please provide description which reflects these issues. The new description shouldn\'t be longer than 100 words.\\

\{format\_instructions\}
            }
        }
    \end{minipage}
    \caption{Prompt for automatic tool description fixing, based on both the tool's description and tool's source code.}
    \label{prompt:autofix_src_desc}
\end{figure}

The file management toolkit benchmark consists of 32 domain-specific environments. The domains for which we have generated file systems are Agriculture and Precision Farming, Biotechnology, Computer Science, Construction and Architecture, Cybersecurity, Data Science, Economics and Market Research, Education and E-Learning, Entertainment and Media, Environmental Science, Finance and Banking, Geology and Geophysics, Government and Public Administration, Graphic Design and Animation, History and Archival Science, Hospitality and Tourism, Human Resources and Recruitment, Journalism and Digital Media, Law and Legal Analytics, Logistics and Supply Chain Management, Machine Learning, Manufacturing and Automation, Meteorology and Climate Science, Music Production and Sound Engineering, Pharmaceuticals and Drug Development, Psychology and Neuroscience, Real Estate, Social Media and Digital Marketing, Sports Science and Analytics, Transportation (Autonomous Vehicles, Traffic Management), Video Game Development, Virtual Reality (VR) and Augmented Reality (AR).

Each task of the benchmark is setup in a Docker container. The Docker container has the initial state of the file system as well as a ReAct agent with the tool under test. The agent is then presented with the task prompt in the initialized environment. Upon completion of the agent's execution, the success of the task is verified using a \texttt{git diff} between the initial and final states (\cref{fig:file_management_diagram}).

For autofixing, GPT-4o is utilized with the following prompts: \Cref{prompt:autofix_us}, \Cref{prompt:autofix_src_desc}, \Cref{prompt:autofix_desc}.

\section{Tested Tools}\label{sec:apdx:tools}
In this section we present the tools used for the evaluation of \tool. The tools are divided into two categories: tools for runtime failure detection and tools for correctness detection.

\subsection{Runtime failure tools}
\begin{figure}[t]
    \centering
    \tiny
    \begin{lstlisting}
class ClothingSorterInput(BaseModel):
    sizes: List[int] = Field(description="The size of the item.")
    designer: str = Field(description="The designer of the item.")  # brand
    colors: List[str] = Field(description="The email of the user who sent the email.")
    min_budget: int = Field(description="The minimum budget you have for this item")
    max_budget: int = Field(description="The maximum budget you have for this item")
    template: str = Field(description="Some template (figures or shapes or items)
                                       that is in search")  # pattern
    collection: str = Field(description="From which collection the item has to be")
    type: str = Field(description="The type of the item 
                                   i.e. clothing piece, shoes, hats etc.")

@valid_prompt(
    "Can you find me T-shirt for at least 10 usd but max 50 usd, I would like it to 
     have one of few patterns i.e. "
    "first square pattern, second traditional bulgarian pattern or just a casual 
     flower pattern, be from the brand "
    "Quick Silver, also from the autumn collection please")
@breaking_prompt("I want to buy shoes in size 23 with price range of 100 to 250 usd.",)
@tool_description("Tool which is underspecified and this leads to 
                   misuse of the parameters of the function")
@tool("find-clothing", return_direct=True)
def find_clothes(sizes: List[int], 
                 designer: str,
                 colors: List[str],
                 min_budget: int, max_budget: int,
                 template: str, collection: str, type: str):
    """
    This tool is useful for filtering clothing items.
    """
    \end{lstlisting}
    \caption{Synthetic tool with under-specified functionality documentation}
    \label{tool:clothing-tool}
\end{figure}

\begin{figure}[t]
    \centering
    \tiny
    \begin{lstlisting}
class ContentInput(BaseModel):
    content: str = Field("The content of a fairy tale which will be summarized")

@breaking_prompt("""Can you shorten this message:
And then, all of a sudden, he said "Pop!" just as if a cork were coming 
out of a bottle.
And Christopher Robin and Rabbit and all Rabbit's friends and relations
 went head-over-heels backwards ... and on the top of them came
Winnie-the-Pooh-free!

So, with a nod of thanks to his friends, he went on with his walk through
the forest, humming proudly to himself. But, Christopher Robin looked
after him lovingly, and said to himself, "Silly old Bear!"
""")
@tool_description(
    "The tool has too concrete description on a common LLM task which 
     will result in the LLM not being able to "
    "recognise when to call it")
@tool('summarize-fairy-tale-content', args_schema=ContentInput, return_direct=True)
def summarize_fairy_tale_content(content: str) -> str:
    """
    Tool to summarize fairy tales given their content. This tool
    can summarize the likes of: Hansel and Gretel,.
    Snow White, The Golden Fish, Rapunzel, The little Mermaid and so on..
    """
    \end{lstlisting}
    \caption{Synthetic tool with over-specified functionality description}
    \label{tool:summarize}
\end{figure}

\begin{figure}[t]
    \centering
    \tiny
    \begin{lstlisting}
class SearchMailInput(BaseModel):
    keywords: str = Field(description="Keywords which will be used for
        filtering (searching) the email inbox.")
    # underspecified, not all enum values are listed for label
    label: str = Field(description="Assigned label to the email.")  
    # labels are -> INBOX, SPAM, TRASH, UNREAD, STARRED, IMPORTANT
    sender_email: str = Field(description="The email of the user who 
                                            sent the email.")


@valid_prompt("What mail did I get in SPAM from royalty@mail.ng about diamonds")
@breaking_prompt("Can you find my email with label finance which is 
related to my taxes?")
@tool_description("Tool which will break because of 
    underspecified fields descriptions for enum values.")
@tool("search-mail", args_schema=SearchMailInput, return_direct=True)
def search_mail(keywords: str, label: str, sender_email: str):
    """
    This tools can search your email box. Useful when you want to 
    find something from your emails!
    """
    \end{lstlisting}
    \caption{Synthetic tool with under-specified arguments description}
    \label{tool:mail}
\end{figure}

The following is a list of tools tested for runtime failures.
Synthetic tools created by the authors of this work:
\texttt{check-basketball-match-stats}, \texttt{personal\_computer\_system\_builder}, \texttt{package-items (synthetic)}, \texttt{compute-bayes-theorem-posteriori}, \texttt{summarise-fairy-tale-content}, \texttt{compute-law-of-cosines}, \texttt{ivanovices\_open\_close\_solver}, \texttt{Jira-issue-work}, \texttt{paper-searcher}, \texttt{json-validator}, \texttt{y\_combinator\_news\_filter}, \texttt{forum-search}, \texttt{car-parts-filter}, \texttt{open-street-map-route-distance}, \texttt{website-searcher}, \texttt{personal-trainer}, \texttt{vans-database-query-engine}, \texttt{search-mail}, \texttt{daily-news-search}, \texttt{json-filter}, \texttt{read-files-from-system}, \texttt{food-delivery-tool}, \texttt{search-documents}, \texttt{Knowledge-base-search}, \texttt{beverage-factory-order}, \texttt{stock-market-report}, \texttt{account-creator}, \texttt{translate-paragraph}, \texttt{search-news-topics}, \texttt{append-to-file}, \texttt{file\_reader}, \texttt{find-clothing}.

The source code for three of these tools can be found in \cref{tool:clothing-tool}, \cref{tool:summarize}, \cref{tool:mail}. These tools are created to showcase the runtime tool failure; their implementation contains only assert statements to validate the function argument constrains, which should be reflected in the tool's documentation.

Tools from the LangChain community \citep{Chase_LangChain_2022}:
\texttt{duckdcukgo\_results\_json}, \texttt{jira-account-creator}, \texttt{arxiv}, \texttt{query\_graphql}, \texttt{ionic\_commerce\_shopping\_tool}, \texttt{pub\_med}, \texttt{Open-street-map-search}, \texttt{request\_delete}, \texttt{copy\_file}, \texttt{duckduckgo\_search}, \texttt{file\_search}, \texttt{request\_patch}, \texttt{youtube\_search}, \texttt{write\_file}, \texttt{list\_directory}, \texttt{read\_file}, \texttt{semanticscholar}, \texttt{Get NASA Image and Video Library video captions location}, \texttt{Wikidata}, \texttt{requests\_put}, \texttt{Get NASA Image and Video Library media metadata location}, \texttt{python\_repl}, \texttt{move\_file}, \texttt{stack\_exchange}, \texttt{requests\_get}, \texttt{requests\_post}, \texttt{file\_delete}.

\subsection{Tools statistics}
\begin{figure}[t]
    \centering
    \includegraphics[width=\linewidth]{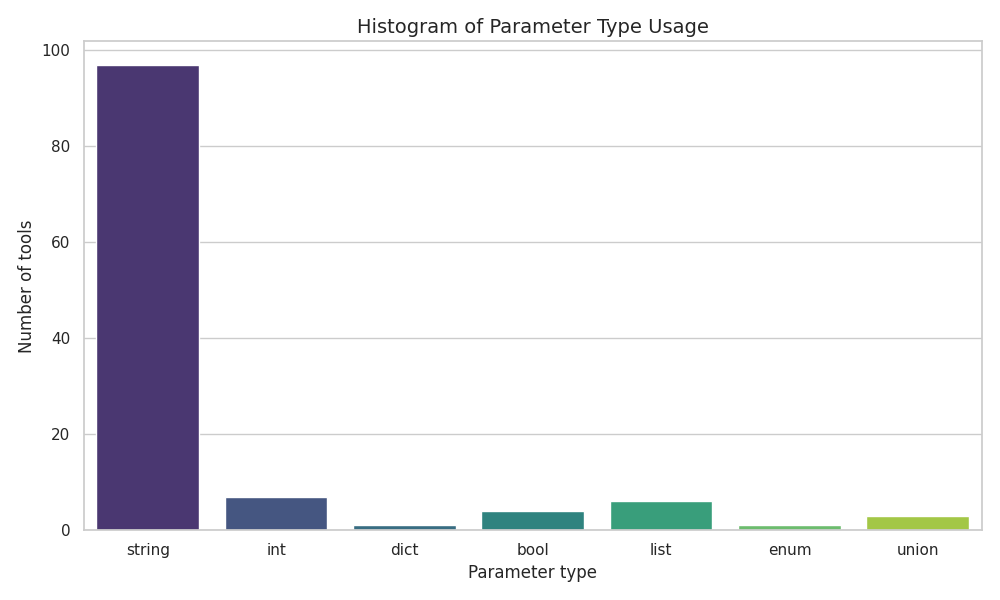}
    \caption{Argument types used by tools}\label{fig:sent_hist}
\end{figure}

\begin{figure}[t]
    \centering
    \includegraphics[width=\linewidth]{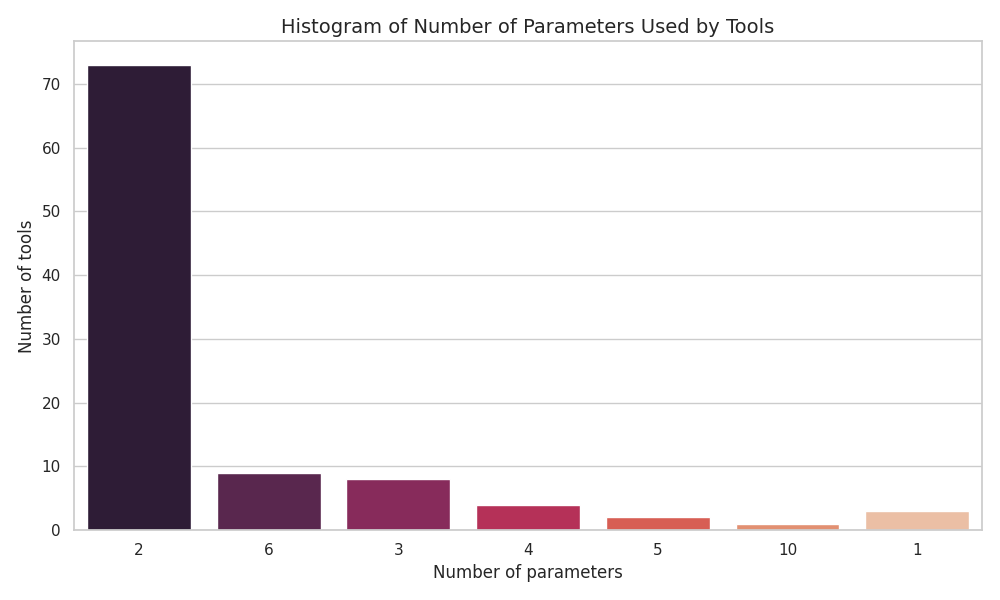}
    \caption{Number of arguments used by tools} \label{fig:num_params}
\end{figure}

\begin{figure}[t]
    \centering
    \tiny
    \begin{minipage}{\textwidth} %
        \fbox{%
            \parbox{\textwidth}{%
               A wrapper around PubMed. Useful for when you need to answer questions about medicine, health, and biomedical topics from biomedical literature, MEDLINE, life science journals, and online books. Input should be a search query.
            }%
        }%
    \end{minipage}

    \caption{PubMed's LangChain tool description}
    \label{fig:pubmed_description}
\end{figure}

\begin{figure}[t]
    \centering
    \tiny
    \begin{minipage}{\textwidth} %
        \fbox{%
            \parbox{\textwidth}{%
              A wrapper around Arxiv.org Useful for when you need to answer questions about Physics, Mathematics, Computer Science, Quantitative Biology, Quantitative Finance, Statistics, Electrical Engineering, and Economics from scientific articles on arxiv.org. Input should be a search query.
            }%
        }%
    \end{minipage}

    \caption{ArXiv's LangChain tool description}
    \label{fig:arxiv_description}
\end{figure}

In this subsection, we have laid down some of the statistics of the Lang chain tools we have collected. That is, the most common argument type is String by a huge margin as per \cref{fig:sent_hist} \cref{fig:num_params} as well as the most common number of arguments is 2, however in Lang chain all of the tools accept a run-time manager in their run method, so in practice most of the tools accept just one argument.

\fi

\clearpage

\end{document}